# Highly efficient reliability analysis of anisotropic heterogeneous slopes: Machine Learning aided Monte Carlo method

Mohammad Aminpour*, PhD, Research Assistant (Corresponding author)

Civil and Infrastructure Engineering Discipline, School of Engineering, Royal Melbourne Institute of Technology (RMIT), Victoria 3001, Australia

E-mail: mohammad.aminpour@rmit.edu.au

Reza Alaie, PhD, Postdoctoral Researcher

Department of Civil Engineering, Faculty of Engineering, University of Guilan, Rasht, Iran

E-mail: alaiereza@phd.guilan.ac.ir

Navid Kardani, PhD, Research Fellow

Civil and Infrastructure Engineering Discipline, School of Engineering, Royal Melbourne Institute of Technology (RMIT), Victoria 3001, Australia

Email: navid.kardani@rmit.edu.au

Sara Moridpour, PhD, Associate Professor

Civil and Infrastructure Engineering Discipline, School of Engineering, Royal Melbourne Institute of Technology (RMIT), Victoria 3001, Australia

Email: sara.moridpour@rmit.edu.au

Majidreza Nazem, PhD, Associate Professor

Civil and Infrastructure Engineering Discipline, School of Engineering, Royal Melbourne Institute of Technology (RMIT), Victoria 3001, Australia

Email: Majidreza.nazem@rmit.edu.au

# Highly efficient reliability analysis of anisotropic heterogeneous slopes: Machine Learning aided Monte Carlo method

Mohammad Aminpour[1*], Reza Alaie[2], Navid Kardani[1], Sara Moridpour[1], Majidreza Nazem[1]

[1] Civil and Infrastructure Engineering, School of Engineering, RMIT University, Melbourne, Australia

[2] Department of Civil Engineering, Faculty of Engineering, University of Guilan, Rasht, Iran.

*Corresponding author:* Mohammad Aminpour, E-mail: mohammad.aminpour@rmit.edu.au

**Abstract:**

Machine Learning (ML) algorithms are increasingly used as surrogate models to increase the efficiency of stochastic reliability analyses in geotechnical engineering. This paper presents a highly efficient ML aided reliability technique that is able to accurately predict the results of a Monte Carlo (MC) reliability study, and yet performs 500 times faster. A complete MC reliability analysis on anisotropic heterogeneous slopes consisting of 120,000 simulated samples is conducted in parallel to the proposed ML aided stochastic technique. Comparing the results of the complete MC study and the proposed ML aided technique, the expected errors of the proposed method are realistically examined. Circumventing the time-consuming computation of factors of safety for the training datasets, the proposed technique is more efficient than previous methods. Different ML models, including Random Forest (RF), Support Vector Machine (SVM) and Artificial Neural Networks (ANN) are presented, optimised and compared. The effects of the size and type of training and testing datasets are discussed. The expected errors of the ML predicted probability of failure are characterised by different levels of soil heterogeneity and anisotropy. Using only 1% of MC samples to train ML surrogate models, the proposed technique can accurately predict the probability of failure with mean errors limited to 0.7%. The proposed technique reduces the computational time required for our study from 306 days to only 14 hours, providing 500 times higher efficiency.

# 1. Introduction

Given the inherent uncertainties in properties of soils and rocks [1], deterministic solutions to typical geotechnical problems have proven insufficient in ensuring a safe design. The inherent soil variability can significantly influence slope stability [2-4]. To address these uncertainties, the application of stochastic methods has progressively become popular amongst geotechnical researchers during the last two decades. Combining the random field theory [5] and the finite element method, the Random Field Finite Element Method (RFEM) is the most widely used stochastic method providing robust probabilistic results on a wide range of geotechnical problems such as bearing capacity [6], consolidation of soft soils [7] and slope stability [8, 9]. However, incorporating the potential variability of mechanical properties of soils using RFEM with stochastic methods such as the Monte Carlo (MC) simulations can be a highly demanding computational approach [10, 11]. Meaningful reliability analyses may involve thousands of random field realisations modelled by numerical methods to create the required stochastic data [12, 13].

When dealing with small values of the probability of failure $p_f$, the MC technique can lead to a high number of required simulations. For small $p_f$ values, the coefficient of variation of the estimated $p_f$ is a function of the number of MC simulations $N_{MC}$ [14]:

$$COV_{p_f} \approx \sqrt{\frac{(1-p_f)}{p_f N_{MC}}} \tag{1}$$

This relationship implies that for a $p_f \ll 1$, to obtain a $COV_{p_f}$ in the order of $O(10^{-1})$, the number of MC simulations $N_{MC}$ should be in the order of $O(100/p_f)$. The high number of required random field finite element simulations can be a significant challenge, demanding remarkable computational time and power. These demands can make it impractical to thoroughly consider the variability properties of geometrically complicated geotechnical problems which in turn, potentially initiates catastrophic risks in geotechnical design, for example in the design of tunnels [15], slopes [16], and foundations [17].

To overcome the computational challenges of RFEM using MC simulations, researchers have introduced different techniques which are mainly categorised into two classes: a) the variance reduction methods such as importance sampling [18] and subset simulation [19, 20], and b) approximated response/surrogate/regression functions (also called meta-models) which are introduced over a reduced number of MC simulations [21-28]. However, the first class of solutions has not been always successful. The importance sampling or subset simulations are shown to be unable to necessarily improve the efficiency of MC methods [29]. The other method to deal with the computational challenges of MC simulations is the construction of surrogate models or response surface methods, including polynomial chaotic expansion, support vector regression, Kriging model, and multiple adaptive regression spline [21-28].

Recently, the machine learning (ML) algorithms have been shown to be efficient response functions trained on reduced MC sample datasets [24, 30-41]. Researchers have mainly used Artificial Neural Networks (ANN) as the ML algorithm representing the response function [30-32]. The method appeared to be hundreds of times more efficient than a complete typical RFEM study with reduced time and power needed [32].

Testing their ML models on limited MC samples, previous studies have not directly validated the outcome of the Machine Learning aided Monte Carlo (MLAMC) analyses in comparison with complete MC reliability results [31, 32]. In a recent publication, He *et al.* [30] presented an efficient deep learning model for predicting soil bearing capacity, trained on various datasets covering a wide range of soil properties and spatial variability coefficients. However, the variation of the model performances (accuracies) was not explicitly shown for each level of heterogeneity (coefficient of variations) and anisotropy (correlation distance) of random fields. Moreover, previous studies utilised simulation datasets with calculated factors of safety (FOS) which are tedious computational tasks. However, a direct MC method can provide the probability of failure without a need for the calculated FOS in each simulation [14]. Thus, when using the surrogate models, reducing the computational work to the determination of failure or non-failure status of slopes can be a significant step in increasing the efficiency of the methods.

In a recent publication, authors systematically explored the efficiency of different machine learning algorithms as surrogate models for MC datasets of anisotropic heterogeneous slopes [42]. Bagging ensemble, Random Forest and Support Vector Classifiers were shown to be the superior models for this problem. The current study will address the implementation of the Machine Learning aided Monte Carlo reliability analysis (MLAMC) on spatially variable random fields. To directly validate this method, the results of the MLAMC results using ML surrogate models trained on limited data are compared with complete stochastic MC reliability analyses obtained from 120,000 MC simulations. Thus, the study realistically evaluates the accuracy of the MLAMC reliability method for different levels of heterogeneity and anisotropy with comparison to the actual MC results. Furthermore, the ML surrogate models in this study are trained on MC datasets where no FOS is calculated, but the failure or non-failure status of random filed slopes are determined. Therefore, the efficiency of ML surrogate models is assessed on a more efficient dataset where the computational time is significantly decreased avoiding FOS calculations. Ultimately, this work characterises the accuracy of the reliability (or the probability of failure) results predicted by ML surrogate models. The expected errors in reliability predictions of ML surrogate models will be determined.

The paper provides the development and evaluation of the MLAMC reliability method as outlined in the following sections. In Section 2, the spatial variability of soil strength parameters is explained with a review of the literature concerning the characteristics of soil variability, including coefficient of variations and scales of fluctuation. Section 3 addresses the method of generating random filed realisations, the details of the random finite difference model and the generated MC simulation datasets. In Section 4, the approach proposed in this study as the MLAMC reliability analysis, and the evaluation process is summarised. In Section 5 the machine learning algorithms used in this study are briefly introduced. Section 6 provides a summary of the optimisation process of ML models using hyperparameter tuning. In Section 7, the performance of MLAMC analyses and the expected errors of $p_f$ predictions are further discussed, concerning the levels of anisotropy and heterogeneity, the size of the train and test datasets and the general versus stochastically specific training samples. Section 8 outlines a summary of the findings and portraits the future research proposals.

## 2. Spatial variability of soil strength parameters

Soil mechanical parameters, although assumed to be constant for a layer of soil in traditional geotechnics, are known to be spatially variable. This variability can be expressed as the combination of the variation trend at a spatial point and the residual variability about the trend. In this study, we assume a stationary random field in which no trend of variations is considered; however, the spatial variability of the strength parameter is characterised as a log-normal distribution. We also consider the anisotropy of the heterogeneity, i.e., the soil strengths can be differently correlated in different dimensions.

To characterise the soil random fields, the point and spatial statistics are chosen following practical ranges recommended in the literature. The point statistics consist of the mean ($\mu$) and the standard deviation ($\sigma$) of the undrained shear strength. The latter can also be introduced as the coefficient of variation, $COV = \sigma/\mu$.

The uncertainties underlying a geotechnical design parameter can root in three primary sources: natural (inherent) variability, measurement error and transformation uncertainty [43]. While the natural uncertainties are related to the geological process of production and modification of soils and rocks, the measurement errors are induced by equipment, operator, and random testing effects. The third source of uncertainties is also introduced when empirical or other correlation models transform field or laboratory measurements into the desired design parameter. An approximate guideline for the combination of these sources of uncertainties associated with the undrained shear strength parameter for clays is presented in Table 1. According to the guideline, we consider a range of 10-50% for the $COV$ of $C_u$ in the random filed generations in this study.

**Table 1.** Approximate guideline for the coefficients of variation of the undrained shear strength, $C_u$ [43]

| Design property [a] | Test [b] | Soil type | Spatial Avg. $COV$ (%)[c] |
|---|---|---|---|
| $C_u$ (UC) | Direct (lab) | Clay | 10 – 40 |
| $C_u$ (UU) | Direct (lab) | Clay | 7 – 25 |
| $C_u$ (UIUC) | Direct (lab) | Clay | 10 – 30 |
| $C_u$ (field) | VST | Clay | 15 – 50 |
| $C_u$ (UU) | $q_T$ | Clay | 30 – 35 |
| $C_u$ (UIUC) | $q_T$ | Clay | 35 - 40 |
| $C_u$ (UU) | N | Clay | 40 – 55 |
| $C_u$ [d] | $K_D$ | Clay | 30 – 55 |

[a] UU: unconsolidated undrained triaxial compression test; UC: unconfined compression test; CIUC: consolidated isotropic undrained triaxial compression test; $C_u$(field): corrected $C_u$ from vane shear test.
[b] VST: vane shear test; $q_T$: corrected cone tip resistance; N: standard penetration test blow count; $K_D$: dilatometer horizontal stress index.
[c] Averaging over 5 m.
[d] Mixture of $C_u$ from UU, UC and VST.

In addition to the point statistics, spatial statistics are involved in the generation of random fields. The spatial statistics are characterised by a critical statistical parameter, namely the *correlation distance*, also called the *scale of fluctuation*, $\delta$. The correlation distance indicates the distance within which the property values show a relatively strong correlation. A very short correlation distance leads to independent property values at different points, which is a rare case in geotechnical profiles. A very large correlation distance will also indicate highly correlated or nearly equal property values at different points. A reasonable correlation distance is practically important in simulating the realistic spatial variability of soil. The correlation distance of soil parameters is shown to be highly different in horizontal and vertical directions. The ratio of the horizontal to vertical correlation distances, $\xi = \frac{\delta h}{\delta v}$ which is the ratio of anisotropy for the soil heterogeneity, is called here the *anisotropy ratio*. Table 2 summarises the values recommended for the correlation distances in the literature. As shown in the table, $\delta_h$ is varied in much larger extent than $\delta_v$. Accordingly, we select a range of 1-25 m for $\delta_h$ and keep an average of $\delta_v = 1m$ for our random field generations.

**Table 2**. Scales of fluctuation (correlation distance) of soil properties.

| Property/Test | Soil type | Horizontal correlation distance, $\delta_h$ (m) | | Vertical correlation distance, $\delta_v$ (m) | | Reference |
|---|---|---|---|---|---|---|
| | | Range | Mean | Range | Mean | |
| $C_u$ (Lab test) | Clay | - | - | 0.8 – 6.1 | 2.5 | [43] |
| $C_u$ (VST)[a] | Clay | 46.0 – 60.0 | 50.7 | 2.0 – 6.2 | 3.8 | [43] |
| Various | Clay | 0.14 – 163.8 | 31.9 | 0.05 – 3.62 | 1.29 | [43] |
| Various | Marine clay | 8.37 – 66 | 30.9 | 0.11 – 6.1 | 1.55 | [43] |
| Various | Sensitive clay | - | - | 1.1 – 2.0 | 1.55 | [43] |
| Various | Silty clay | 9.65 – 45.4 | 29.8 | 0.095 – 6.47 | 1.40 | [43] |
| Various | Soft clay | 22.2 – 80 | 47.6 | 0.14 – 6.2 | 1.70 | [43] |
| CPT | Clay | - | - | 0.63 – 2.55 | 1.61 | [44] |
| CPT | Clay | 23- 66 | 44.5 | 0.1 – 2.2 | 0.9 | [45] |
| CPT | Sand, clay | 3 – 80 | 47.9 | 0.2 – 0.5 | 0.3 | [45] |
| CPT | Clay | - | - | 0.19 – 0.72 | 0.36 | [46] |
| VST | Clay | 10 – 40 | - | 1 – 3 | - | [47] |
| CPT | Sand, clay | - | - | 0.13 – 1.11 | 0.63 | [48] |
| CPT | Sand, clay, silt | 66 – 1546 | - | 1.72 – 2.53 | 2.02 | [49] |
| Cohesion | Clay | 10 - 40 | - | 0.5 – 3 | - | [50] |
| CPTu | Sand, clay | 12.15 – 16.11 | - | 0.07 – 1.32 | 0.3 | [51] |

[a] Vane shear test

## 3. Random field Monte Carlo Simulations

### 3.1 Random field generation

A random field is defined as a random function over an arbitrary domain. Statistical parameters to characterise the random field are the mean value $\mu$, standard deviation $\sigma$ and correlation distance $\delta$. The latter characterises how rapidly the field varies in space and is captured by the covariance function. The random field is known as stationary when the complete probability distribution is independent of absolute location, depending only on vector separation (i.e., distance and direction) between two spatial points in the field. In contrast, a non-stationary field is statistically heterogeneous. For example, the mean of the random parameter can be a function of location, such as changing with a trend with depth [52]. Assuming a Gaussian and stationary random field, we take the undrained shear strength parameter, $C_u$, as the random variable.

Random fields can be generated using several methods, including Covariance Matrix Decomposition, Moving Average, Fast Fourier Transform, Turning Bands Method and Local Average Subdivision Method [53]. Here we use the Cholesky decomposition technique developed in [54] and utilised in various geotechnical studies [55, 56].

In this study, the random field is defined as [57]

$$C_u(\tilde{x}) = exp(L.\varepsilon + \mu_{lnC_u(\tilde{x})}) \qquad (2)$$

where $C_u(\tilde{x})$ is the undrained shear strength of soil at the spatial position $\tilde{x}$, $\varepsilon$ is an independent normally distributed random variable with zero mean and unit variance, $\mu_{lnC_u(\tilde{x})}$ is the mean of the logarithm of $C_u$, and L is a lower-triangular matrix computed from the decomposition of the covariance matrix using the Cholesky decomposition technique [53]. The anisotropic exponential Markovian covariance function is defined as [1]

$$A(l_x, l_y) = \sigma^2_{ln\,C_u} exp\left(-\frac{|l_x|}{\delta_h} - \frac{|l_y|}{\delta_v}\right) \qquad (3)$$

where $l_x$ and $l_y$ are the horizontal and vertical distances between two points, $\delta_h$ and $\delta_v$ are the horizontal and vertical correlation distances, respectively, and $\sigma_{ln\,C_u}$ is the standard deviation of the logarithms of undrained shear strength.

In the Cholesky decomposition technique, the covariance between the logarithms of the random variable values (here the undrained shear strength) at any two points is decomposed into

$$A = LL^T \qquad (4)$$

where, $T$ stands for transpose.

The mean ($\mu$) and the standard deviation ($\sigma$) of the logarithms of $C_u$ are given as

$$\mu_{ln\,C_u} = ln\,\mu_{C_u} - \frac{1}{2}\sigma^2_{ln\,C_u} \qquad (5)$$

$$\sigma_{lnC_u} = \sqrt{ln(1 + COV^2_{C_u})} \qquad (6)$$

where, $COV_{C_u}$ is the coefficient of variation of $C_u$ and is obtained by

$$COV_{C_u} = \frac{\sigma_{C_u}}{\mu_{C_u}}. \qquad (7)$$

The mean of the undrained shear strength, $\mu_{C_u}$ in our simulations varies from 18.6 to 33.5 kPa. These $\mu_{C_u}$ values are selected based on the factors of safety (FOS) associated with homogeneous slopes where the shear strength of soil is equal to $\mu_{C_u}$ values. The FOS associated with a homogeneous slope with $C_u = 18.6$ kPa is equal to 1. The other $\mu_{C_u}$ values are selected as 1.2, 1.4, 1.6 and 1.8 times of 18.6 kPa. The coefficient of variations is changed as 0.1, 0.3 and 0.5. The vertical correlation distance, $\delta_v$ is kept as 1 m, whereas the horizontal correlation distance, $\delta_h$ is varied from 1 to 25 m. The summary of statistical parameters used in this study is shown in Table 3.

Table 3. The statistical parameters of spatial variability of soil used in this study

| Parameter | Values |
|---|---|
| Mean of undrained shear strength, $\mu_{C_u}$ (kPa) | 18.6, 22.3, 26, 29.7, 33.5 |
| Coefficient of variation, $COV$ | 0.1, 0.3, 0.5 |
| Horizontal correlation distance, $\delta_h$, m | 1, 6, 12, 25 |
| Vertical correlation distance, $\delta_v$, m | 1 |

## 3.2 Finite difference random field model

A typical slope geometry is considered for this study with a slope angle of 45°, slope height of 5 m and foundation depth of 10 m from the top of the slope, as shown in Fig. 1. A rigid boundary surface is adapted at the foundation level where all movements are restricted simulating hard rock beds. The horizontal displacements are also restricted at the vertical boundaries of the model. A linear elastic-perfectly plastic stress-strain behaviour is defined for the soil incorporating the Mohr-Coulomb failure criterion. The shear modulus of soil $G$ is assumed to be a linear function of the undrained shear strength as $G = I_R C_u$ where $I_R$ is the rigidity index defined as the ratio between the shear modulus G and undrained shear strength. $I_R = 800$ denoting a moderately stiff soil is assumed per ranges examined in the literature (e.g., $I_R = 300 - 1500$ [58]). The soil unit weight $\gamma_{sat} = 20 \text{kN/m}^3$ is held constant in all simulations.

The analyses are performed using FLAC (*FLAC Itasca*), where the mesh size was examined to be sufficiently fine in acquiring accurate results with less than 5% relative errors compared to highly fine and computationally expensive mesh size. A four-nodded quadrilateral grid with a 0.5m side dimension was adopted for this study resulting in 800 cells which are attributed to a different $C_u$ value in a random field realisation (Fig. 1).

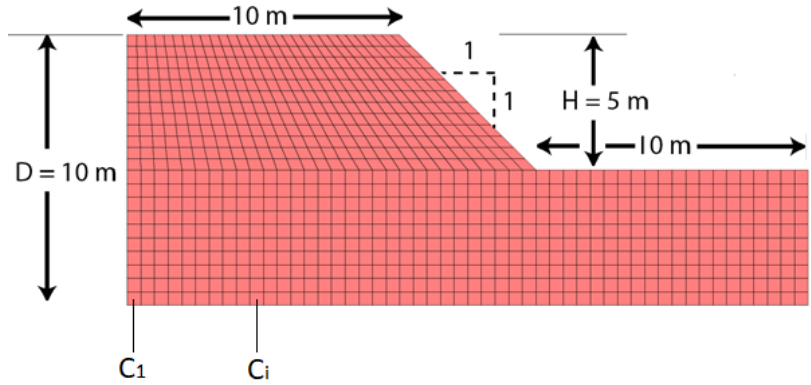

**h_i**: a single random filed sample  $h_i = \{C_1, C_2, C_3, ..., C_n\}$
**H**: entire random filed data  $H = \{h_1, h_2, h_3, ..., h_m\}$
**L**: label vector (failed or stable)  $L = \{l_1, l_2, l_3, ..., l_m\}$
$f(H) = L$   f: ML surrogate model

**Fig. 1** Slope geometry for the random field study showing Random Finite Difference Model comprising 800 random variable cells ($C_i$). Each cell attributes to a random strength variable. All variables together create the vector of features $h_i$ for each sample used to train the ML surrogate models.

The strength reduction technique, which is normally used to obtain the factor of safety of slopes in numerical simulations, is computationally demanding with many iterations required to assess a single stability analysis. To identify if the slope is stable or not, an alternative criterion for slope failure is adapted where a combined criterion for plastic zones and the velocity of grids is assessed in the finite difference model. The failure surface is detected if a contiguous region of active plastic zones connecting two boundary surfaces is developed. As a velocity criterion, the failure is also defined when both the velocity gradient and amplitude are greater than determined thresholds. This alternative failure criterion is proved efficient with accurate results as compared to stochastic stability results obtained from the strength reduction technique (see Ref. [59] for more details).

### 3.3 Monte Carlo reliability analysis

Monte Carlo reliability analysis involves the simulation of a sufficient number of samples for desired statistical parameters by which the statistical results such as the probability of failure converge to constant values. For this study, it can be shown that 2000 samples are sufficient [59]. This number aligns with findings suggesting that a typical number of 2500 simulations can provide reasonable precision and reproducibility for random field finite element studies [53]. Thus, for each set of statistical parameters ($\mu_{C_u}, \sigma_{C_u}, \delta_h, \delta_v$), 2000 random field realisations are generated, each simulated using the finite difference method to determine the stability conditions of slopes. The results of all analyses provide the Monte Carlo data for the reliability analysis. The details of the random fields generated in this study are summarised in Table 4.

Table 4. The characteristics of the random fields generated for slope stability Monte Carlo reliability analysis in this study

| Random field | COV | $\delta_v$ (m) | $\xi$ | Mean of undrained shear strength, $\mu_{C_u}$ (kPa) | No. of Monte Carlo simulations |
|---|---|---|---|---|---|
| $U_{0.1,1}$ | 0.1 | 1 | 1 | 18.6, 22.3, 26, 29.7, 33.5 | 5 × 2000 |
| $U_{0.1,6}$ | 0.1 | 1 | 6 | 18.6, 22.3, 26, 29.7, 33.5 | 5 × 2000 |
| $U_{0.1,12}$ | 0.1 | 1 | 12 | 18.6, 22.3, 26, 29.7, 33.5 | 5 × 2000 |
| $U_{0.1,25}$ | 0.1 | 1 | 25 | 18.6, 22.3, 26, 29.7, 33.5 | 5 × 2000 |
| $U_{0.3,1}$ | 0.3 | 1 | 1 | 18.6, 22.3, 26, 29.7, 33.5 | 5 × 2000 |
| $U_{0.3,6}$ | 0.3 | 1 | 6 | 18.6, 22.3, 26, 29.7, 33.5 | 5 × 2000 |
| $U_{0.3,12}$ | 0.3 | 1 | 12 | 18.6, 22.3, 26, 29.7, 33.5 | 5 × 2000 |
| $U_{0.3,25}$ | 0.3 | 1 | 25 | 18.6, 22.3, 26, 29.7, 33.5 | 5 × 2000 |
| $U_{0.5,1}$ | 0.5 | 1 | 1 | 18.6, 22.3, 26, 29.7, 33.5 | 5 × 2000 |
| $U_{0.5,6}$ | 0.5 | 1 | 6 | 18.6, 22.3, 26, 29.7, 33.5 | 5 × 2000 |
| $U_{0.5,12}$ | 0.5 | 1 | 12 | 18.6, 22.3, 26, 29.7, 33.5 | 5 × 2000 |
| $U_{0.5,25}$ | 0.5 | 1 | 25 | 18.6, 22.3, 26, 29.7, 33.5 | 5 × 2000 |
| | | | | total | 120000 |

The probability of failure, $p_f$ is defined as the probability of the slope being unstable for each case of statistical parameters. In an MC analysis, we can define the $p_f$ as the number of simulations leading to failure divided by the total number of realisations, which is equal to

$$p_f = (N_f/N) \times 100 \qquad (8)$$

where, the probability of failure, $p_f$ is in percentage, $N_f$ is the number of simulations resulting in failure and $N$ is the total number of random field realisations associated with a certain set of statistical parameters.

In Fig. , the random spatial variability of the strength parameter in simulated slopes is demonstrated. Each sub-figure in Fig. visualises a single example of the random fields from the set of 2000 cases generated for each set of stochastic parameters ($\mu_{C_u}$, $COV$ and $\xi$). The finite difference model of the slopes is discretised to 800 cells with a random $C_u$ value. The probability distribution functions of $C_u$ values are shown in the right panels of Fig. . The insets in the right panels demonstrate the PDFs of the $ln(C_u/\mu_{C_u})$ values illustrating the log-normal distribution of data.

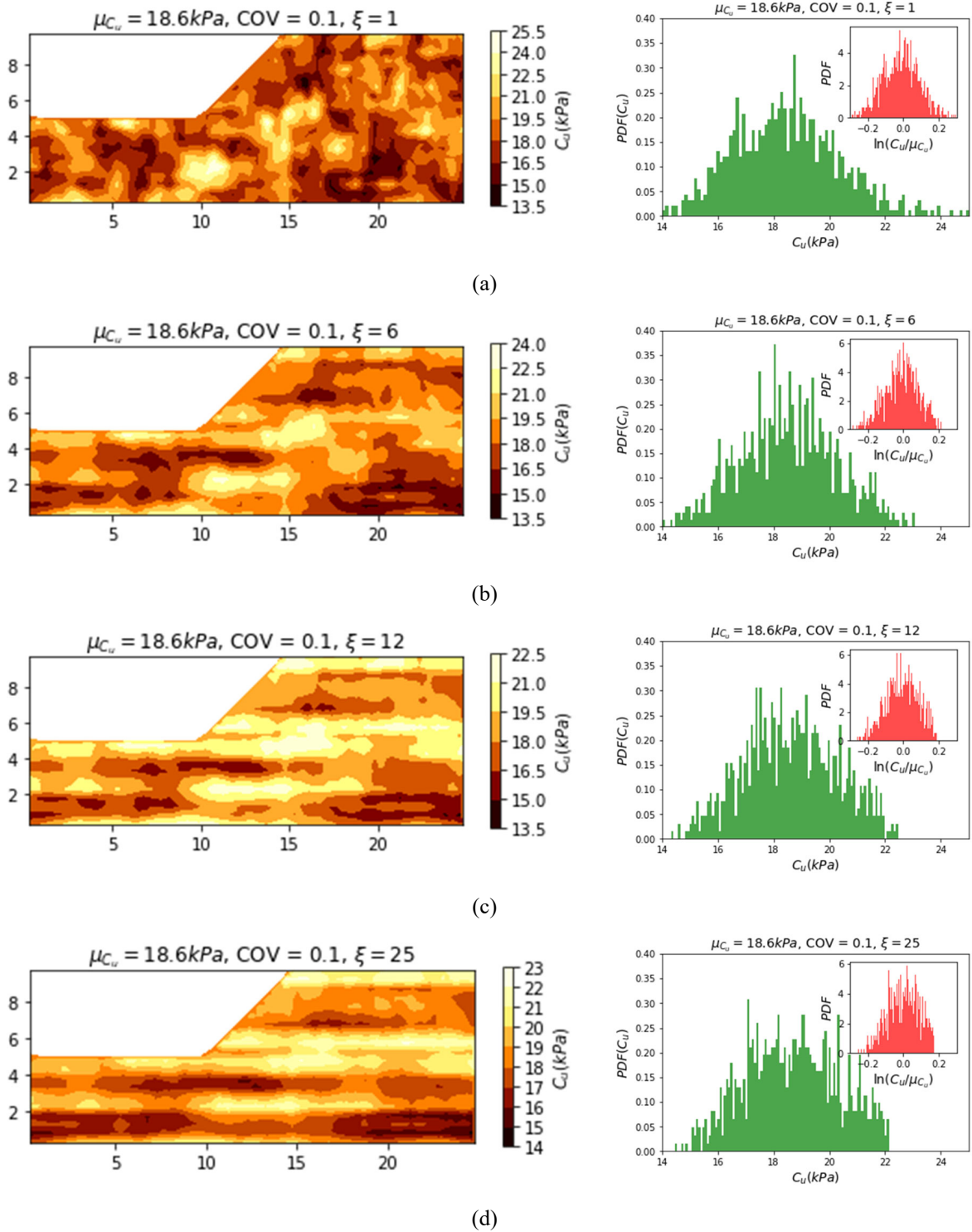

**Fig. 2** Random field spatial variability of undrained shear strength of the soil. Left panels are the contour maps of the random fields and the right panels are the corresponding probability distribution function of the strength parameter ($C_u$). Plots correspond to $COV = 0.1$ with $\xi = 1$ (a), 6 (b), 12 (c) and 25 (d), and $COV = 0.5$ with $\xi = 1$ (e), 6 (f), 12 (g) and 25 (h). The mean value of $C_u$ for all illustrated slopes here is 18.6 kPa.

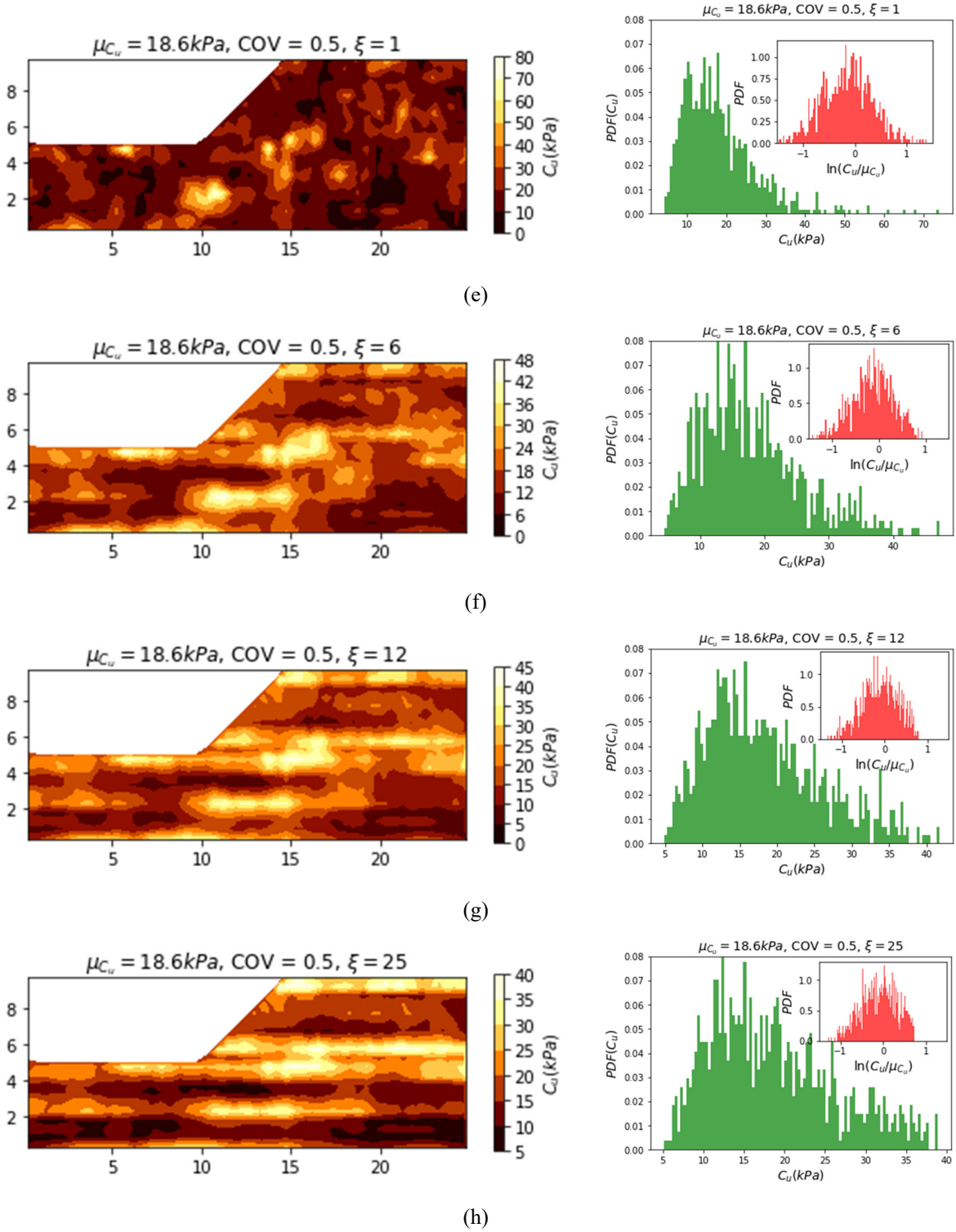

(e)

(f)

(g)

(h)

**Fig. 2** Continued.

## 4. Machine Learning Aided Monte Carlo (MLAMC) Analysis

In Machine Learning Aided Monte Carlo (MLAMC) analysis, ML algorithms are used as surrogate models. A limited number of random field simulations (1-2% of the complete MC dataset) is conducted using numerical methods. The random field spatial variables are the input parameters for the surrogate models. Here, the distribution of $C_u$ variable throughout 800 cells of the finite difference model creates the input vector of features for the surrogate models. The output of these models is the response of numerical simulations, such as FOS or the stability status. Here, the binary status of slopes as failed or stable is the output. Surrogate models are trained and tested using the data obtained from the simulated samples. If the outcome of the surrogate models (e.g., probability of failure) is accurate enough with errors in acceptable ranges, the surrogate models are applied on the entire random filed data to predict the outcome of numerical simulations for the whole MC dataset. Therefore, instead of full-sized MC datasets (normally in the order of ~$10^3$-$10^5$ simulations), only a limited number of random filed simulations in scales of tens to a couple of hundreds is required [30-34]. Fig. 3 shows a flowchart summarising the procedure of the proposed MLAMC reliability analysis.

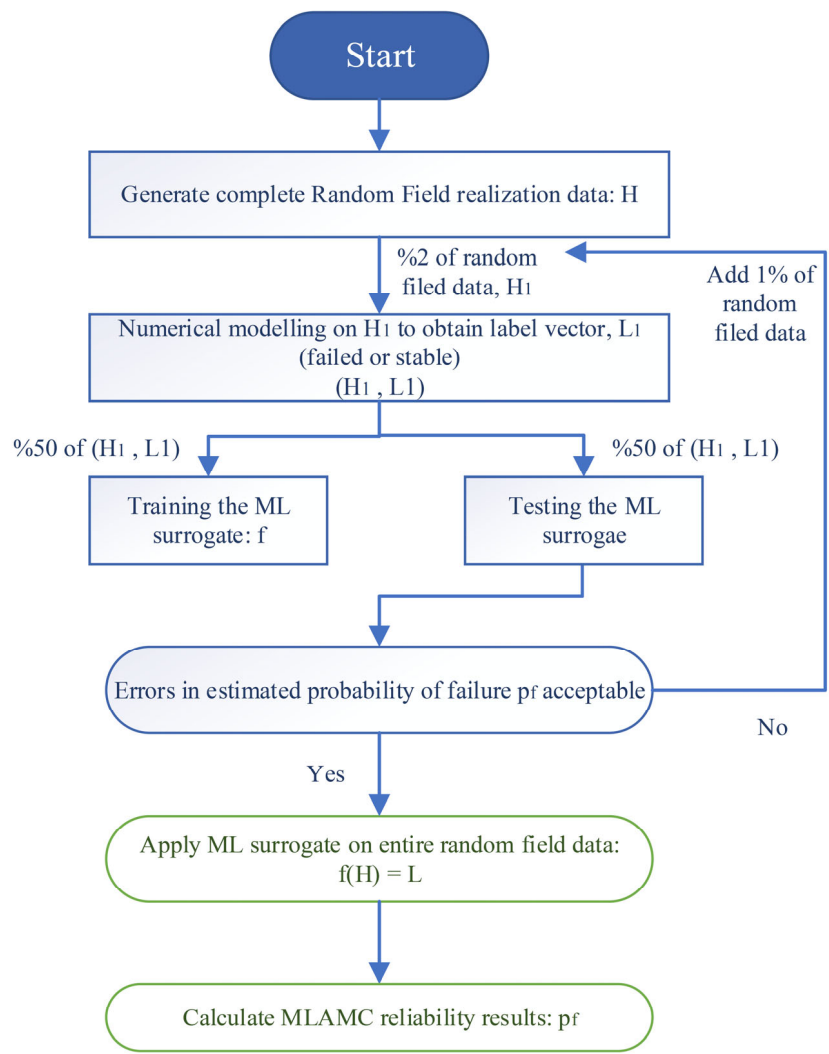

**Fig. 3** The proposed procedure of using ML surrogate models for MLAMC reliability analysis

In this study, a full-sized MC reliability analysis, as well as the corresponding MLAMC analysis are conducted. Comparing the results of both methods, this study accurately evaluates the results of the MLAMC method providing the scales of expected errors in predicted reliability values. A graphical summary of the approach implemented in this paper is demonstrated in Fig. 4.

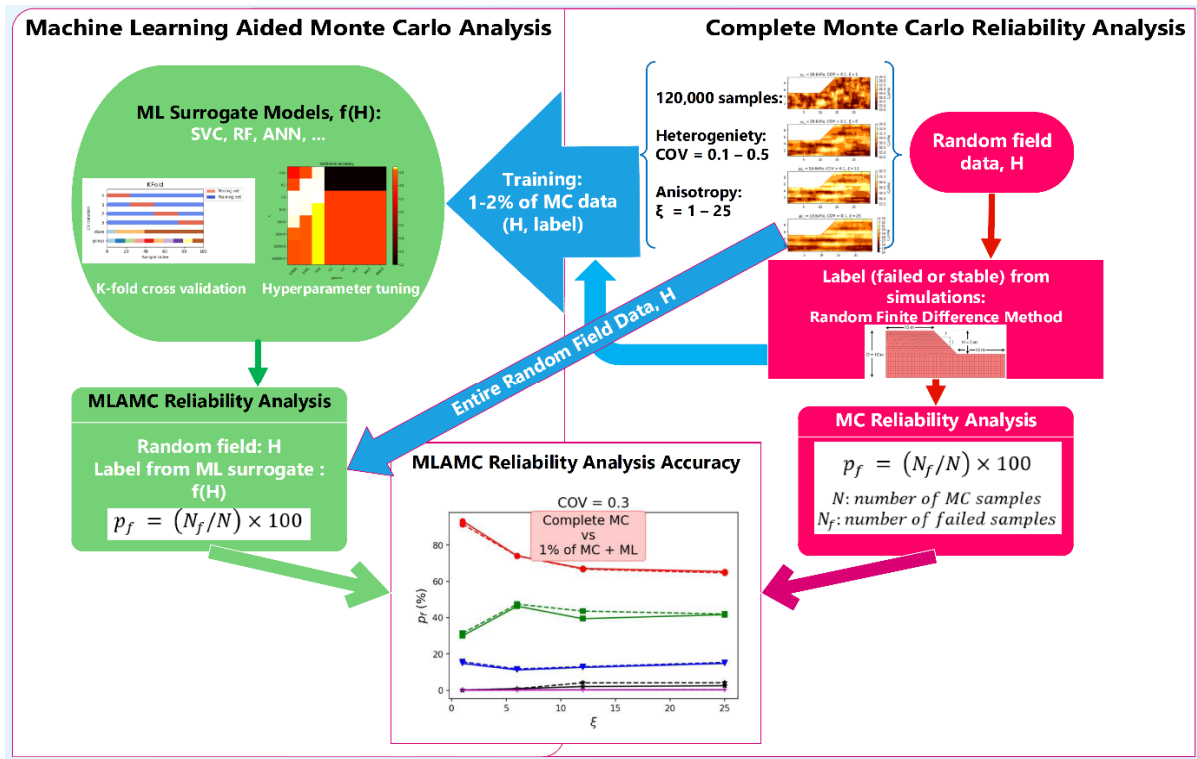

**Fig. 4** A graphical summary of the approach in evaluating MLAMC reliability analysis in comparison with a complete Monte Carlo analysis in this study

## 5. Machine Learning Algorithms

### 5.1. Support Vector Classifier

As a supervised learning technique, Support Vector Machine (SVM) is often utilised for three primary objectives: classification, regression and pattern recognition. The SVM used for classification is called a Support Vector Classifier (SVC). Two significant characteristics distinguishing SVM from basic ANN are fast computation to achieve globally optimal and effective performance to resist overfitting. Given a dataset with N samples and a represented result type of y (1 or 0), the following training vectors can be used for binary classification:

$$D = (x_1, y_1), (x_2, y_2), \ldots, (x_N, y_N); \; x \in R^n, \; y_i \in {0, 1} \qquad (9)$$

where, *n* influencing features exist in the n-dimensional space that may address the decision boundary. Soft-margin SVMs impose penalties and can perform well for nonlinear classification when combined with the kernel technique. Therefore, every function *f(x)* in the SVM kernel trick may be shown in the following form [60]:

$$f(x) = w^T \phi(x) + b = 0. \qquad (10)$$

$W^T$ is the transposed vector of the output layer, where b denotes the bias. The input variable *x* is given as a $N \times n$ matrix, and $\phi(x)$ is a kernel function.

The following problem of reduction is proposed to determine w and b [61]:

Minimise: $\frac{1}{2} w^T + C \sum_{i=1}^{n} \chi_i$ \qquad (11)

Subjected to: $y_i(w. x_i + b) \geq 1 - \chi_i$

Here, $C$ is a penalty coefficient while $\chi_i \geq 0$ denotes slack variables which relate to the misclassification effects.

## 5.2. Random Forest (RF)

Random Forest (RF) is a method for ensemble learning which is based on divide-and-conquer, a popular technique in real-world classification and regression. RF employs bootstrap aggregation, commonly known as bagging, to create an ensemble of randomly generated base learners (unpruned DTs). The core element of RF is the creation of a collection of DTs subjected to controlled alternation. Another critical stage of the training process is the random choice of features. The method creates a subset of M important variables for each node in the trees. A variety of influencing factors is chosen to guarantee DT variation. As a result, each of the DTs independently assesses the forest. The results are obtained by voting on the whole tree. In general, the RF algorithm includes the following four steps:

Step 1: Samples are randomly chosen from a dataset.

Step 2: For each sample, a decision tree is constructed, and prediction results are obtained from the decision tree.

Step 3: Voting is performed on the outcome of each forecast.

Step 4: The final prediction is assigned to the most chosen prediction result.

Additional information regarding RF and its usage in slope stability predictions may be found in references [62, 63].

### 5.3. Artificial Neural Networks (ANN)

ANNs are inspired by the structure of the human brain and consist of basic processing units referred to as artificial nodes or neurons that may be utilised in computational techniques. This structure enables ANNs to be used effectively for various mathematical tasks, such as pattern identification and function estimation. An ANN is composed of three primary layers in which neurons are located. Neurons in the inputs or first layer have the same input variables as others in the outputs or final layer. Between these two levels is at minimum one hidden layer. The input layer transmits the signal, the hidden layers act as the network's computational engine, and the output layer generates predictions depending on the inputs. As the primary parameters of an ANN, weights and bias define the network's degree of freedom and the connections between linked neurons in a layer, respectively. Excluding the input nodes, each node determines its output using a nonlinear activation transfer function while presenting several inputs. The following equations describe the most frequently used activation functions:

Sigmoid function: $f(z) = \frac{1}{1+e^{-z}}$ (12)

Rectified Linear Unit function (ReLU): $f(z) = max(0, z)$ (13)

Hyperbolic tangent function: $f(z) = \frac{e^z - e^{-z}}{e^z + e^{-z}}$ (14)

The calculated output is then used as an input for the subsequent node, and this is continued until a dependable solution to the initial problem is discovered. The backpropagation method is used to compute the error associated with comparing the observed result (i.e., the problem goal) to the projected outcome (i.e., the network's outcome). This inaccuracy is transmitted back through the ANN structure one layer at a time, and the weights are modified according to their contributions to the error.

The algorithms used to change the attributes of ANNs such as weights and learning rate to reduce the losses are known as optimisers. Optimisers are used to solve optimisation problems by minimising the function. The Stochastic Gradient Descent (*SGD*) and Adaptive Moment Estimation (*ADAM*) are two of the most widely used optimisers [64, 65].

### 6. Hyperparameter optimisation

According to the literature, SVC and RF are among the best performing classifiers for random filed slope stability predictions [42]. These classifiers are selected in this study for further evaluation of MLAMC analysis. Additionally, the performance of ANN as a deep learning algorithm which is normally used in similar studies [30-32] is evaluated for this problem. A grid search k-fold cross-validation (CV) technique is employed to identify the extent of the dependency of the performance levels on the variation of hyperparameters. By searching over a range of hyperparameters, the method identifies the optimised parameters for each ML algorithm. Using a 5-fold CV scheme, the sensitivity of performance values is assessed. The CV scheme results in a distribution of performance values where the mean values are reported here. We consider the optimisation individually for different datasets, from low to high levels of heterogeneity and anisotropy.

The following hyperparameters are the major tuning parameters for the RF model: *n_estimators* as the number of trees in the forest; *max_depth* as the maximum depth of the tree, *max_features* as the number of features to consider when looking for the best split, which is a function of *n_features* (sqrt or log2), and *criterion* as the function (*Gini* or *entropy*) to measure the quality of a split. Through the optimisation process, the RF model is found to be slightly sensitive to the choice of hyperparameters with less than 2.5% of variations in the obtained classification accuracy.

Despite the slight sensitivity of the RF model on hyperparameter tuning, the SVC performance appears to be critically dependent on the choice of hyperparameters. The right choice of *c* and *gamma* parameters can change the classification accuracies by up to 70%. A combination of *c* = 1.0 and *gamma* = 0.001 can be selected as the optimised hyperparameters common for all datasets.

The choice of hyperparameters is shown to be moderately impactful on the performance of ANNs in this study. The model optimiser method (*ADAM* or stochastic gradient descent, *GSD*), *L2 regularisation*, *dropout rate* and the *number of units (neurons)* are the most important hyperparameters for ANNs optimised in this study. The classifier predictions show improvements by up to 11.6% for the range of hyperparameters examined.

While there is no straight rule to determine the optimised number of hidden layers in training an ANN, systematic experimentation to discover the best choices for any specific dataset can be a generally accepted approach. We investigated the effects of hidden layers by incorporating 1, 2, 3 and 6 hidden layers for each of the datasets. The test indicated that an increased number of hidden layers, although rising the computational cost of the model, does not guarantee higher performance. As an important hyperparameter effective in the performance of ANNs, the choice of *dropout layers* is also examined. Dropout is a regularisation technique invented to prevent a network from overfitting. In most cases of heterogeneity and anisotropy levels, in terms of accuracies, a single hidden layer joined with a single dropout layer outperforms the models with 1 to 6 hidden layers without a dropout layer.

Table 5 summarises the optimisation process of ML models in which the tested hyperparameters and their ranges, the optimised value and the impact of the optimisation on the improvement of classifier accuracies are shown.

Table 5. The summary of hyperparameter optimisation results

| ML model | Dataset | Hyperparameter | Range | Optimised value | Improvement of classification accuracy by hyperparameter optimisation |
|---|---|---|---|---|---|
| RF | $U_{0.1,1}$, $U_{0.5,25}$ | *max_features* | auto, log2, sqrt | log2 | <2.5% |
|  |  | *max_depth* | 4, 10, 100, 1000, 5000 | 1000 |  |
|  |  | *criterion* | Entropy, Gini | Gini |  |
| SVC | All U ($COV$ = 0.1 - 0.5 $\xi = 1 - 25$) | *c* | $10^{-2} - 10^{5}$ | 1.0 | Up to 70% |
|  |  | *gamma* | $10^{-4} - 10^{3}$ | $10^{-3}$ |  |
| ANN | $U_{0.1,1}$, $U_{0.5,25}$ | *optimiser* | ADAM, SGD | SGD | <12% |
|  |  | *L2* | 0.001, 0.1 | 0.1 ($U_{0.1,1}$), 0.001 ($U_{0.5,25}$) |  |
|  |  | *Dropout rate* | 0.1, 0.5 | 0.5 |  |
|  |  | *Units* | 16, 64, 800 | 800 ($U_{0.1,1}$), 64 ($U_{0.5,25}$) |  |

## 7. Results and discussion

### 7.1 The effects of heterogeneity and anisotropy on the classification performances

In this section, the ML surrogate models are evaluated for the classification task predicting the failure or non-failure status of slopes. Therefore, each slope is predicted as failed or stable, regardless of the probability of failure. The RF, SVC and ANN classifiers are examined on datasets with different levels of heterogeneity ($COV$ = 0.1-0.5) and anisotropy ($\xi = 1 - 25$). The probability of failure obtained from the models is discussed in the following sections.

The Accuracy score (ACC) and the area under the Receiver Operating Characteristic (ROC) curve, i.e., AUC score are used as the model performance metrics. The ACC score is defined as

$$\text{Accuracy (ACC)} = \frac{\text{True Positive + True Negative}}{\text{total instances}}. \tag{15}$$

The ROC curve also plots the true positive rate against the false positive rate at various cut-off values. The area under the ROC curve (AUC score) is commonly used as a measure of the performance of classification. AUC ranges in value from 0 to 1, representing a totally wrong to fully correct prediction by a classifier, respectively.

The ML classifier models are trained on 5% of data (500 samples) for each U dataset (see Table 4) where the performance scores shown in Fig. 5 are obtained on the testing datasets which are the remaining 95% of the datasets. A general trend in all models is a declining performance (decreased accuracy and AUC scores) with increasing heterogeneity and anisotropy. The reason for this behaviour may lie in the increased complexity of the failure mechanisms with such a rise in heterogeneity and anisotropy. A less heterogeneous slope is more likely to fail via typical shear surfaces which are better learned by the classifiers. A semi-circular geometry of the failure surface can be expected in less heterogeneous slopes where the instability can be better correlated with the average of shear strength values. With increasing heterogeneity, the predictability of the slope stability decreases as the failure mechanisms can be more complex attributing to the interconnection of local week zones and irregular surfaces of failure.

A higher anisotropy ($\xi$) can lead to lower predictability of the stability status of slopes as evident in Fig. 5. The complexity of the failure mechanisms may increase by higher anisotropy where not only the local week zones (due to heterogeneity) can contribute to the failure, but the week layers distributed spatially into the field are presumably parts of irregular failure surfaces. Thus, the possible failure mechanisms may include any geometry of failure from a circular deep surface to a local shearing zone or a shallow or deep layering slide. With the low repeatability of different mechanisms in limited training datasets, the prediction ability of the models is decreased.

In general, the optimised SVC model looks more appropriate on all anisotropy and heterogeneity levels with performance scores similar to or higher than those of RF and ANN models (Fig. 5). However, the SVC can be highly sensitive to hyperparameters; thus, care should be taken in optimising those parameters. Despite detailed optimisation, a deep learning approach using ANN with one hidden layer plus one dropout layer does not appear to be advantageous compared to the other selected classification models, i.e., RF and SVC.

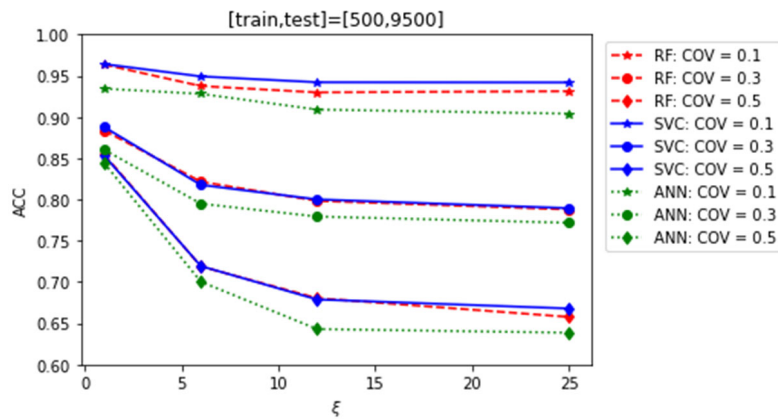

(a)

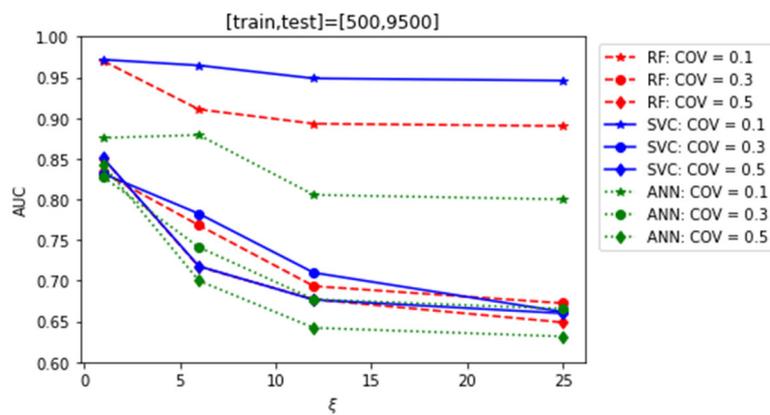

(b)

**Fig. 5** Slope stability classification performances. Accuracy Score, ACC (a) and Area Under Curve, AUC (b) for classification task versus anisotropy ratio $\xi$ on unseen testing data using Random Forest (RF), Support Vector (SVC) and ANN with 1 hidden layer plus 1 dropout layer. Models are trained on 500 samples and tested on 9500 samples.

### 7.2 The effects of the testing dataset size on the classification performances

In previous studies on the application of ML surrogate models, the training and testing of ML models have been normally performed on a limited number of MC simulations [31, 32] where the combination of train and test data for each case of random field statistical parameters has not reached the scales of a full dataset of MC simulation. One of the research questions that our study is aiming to answer is whether the performance metrics of the MLAMC analysis will change if the ML model is tested against the full dataset of MC simulations instead of a limited number of testing samples.

To answer this question, for each set of *COV* and *ξ* values, we first train the RF, SVC and ANN models on a trainset of 500 samples (5% of data) randomly extracted from each set of 10,000 samples. Then the fitted models are tested against two datasets: a) a random test dataset of 500 samples, and b) the entire remaining 9500 samples as the test dataset. Therefore, if the performance metrics remain similar for both a and b test datasets, we can conclude that testing the fitted ML models on only 5% of data could be statistically equivalent to testing against the entire data.

Table 6 summarises the ACC and AUC metrics on the slope stability classification task obtained for RF, SVC and ANN models trained on 500 samples, but tested against a) 500 and b) 9500 samples. The similarity of each pair of metrics obtained for a and b cases is consistent for all random field datasets with the differences of the metric pairs being meaningfully small. The maximum difference of ACC and AUC metrics when the size of testing datasets changes is 2.9% and 3.9%, respectively. The results can be interpreted as the model performances being statistically independent on the size of the test datasets when 5% of an MC full dataset is available for testing the trained models. In other words, in our study, testing the models on 5 or 95% of the whole MC dataset can be almost statistically equivalent. This examination can be a significant step in evaluating the MLAMC method as we can confirm that generating a small portion of the MC dataset can be statistically adequate for training and testing of the models with reliable accuracy levels.

**Table 6** The performance metrics of RF, SVC and ANN models in classifying the slope stability status for different sizes of testing datasets.

| Random field label | Train set size | Test set size | RF | | SVC | | ANN | | Max difference of | |
|---|---|---|---|---|---|---|---|---|---|---|
| | | | ACC | AUC | ACC | AUC | ACC | AUC | ACC | AUC |
| $U_{0.1,1}$ | 500 | 500 | 0.962 | 0.963 | 0.966 | 0.980 | 0.946 | 0.897 | 0.012 | 0.021 |
| | | 9500 | 0.961 | 0.961 | 0.964 | 0.972 | 0.934 | 0.876 | | |
| $U_{0.1,6}$ | 500 | 500 | 0.938 | 0.877 | 0.960 | 0.971 | 0.930 | 0.867 | 0.011 | 0.010 |
| | | 9500 | 0.928 | 0.867 | 0.949 | 0.965 | 0.928 | 0.879 | | |
| $U_{0.1,12}$ | 500 | 500 | 0.940 | 0.874 | 0.958 | 0.965 | 0.936 | 0.845 | 0.027 | 0.039 |
| | | 9500 | 0.926 | 0.871 | 0.942 | 0.949 | 0.909 | 0.806 | | |
| $U_{0.1,25}$ | 500 | 500 | 0.930 | 0.828 | 0.948 | 0.933 | 0.910 | 0.785 | 0.008 | 0.013 |
| | | 9500 | 0.922 | 0.843 | 0.942 | 0.946 | 0.904 | 0.800 | | |
| $U_{0.3,1}$ | 500 | 500 | 0.894 | 0.835 | 0.894 | 0.835 | 0.878 | 0.835 | 0.018 | 0.007 |
| | | 9500 | 0.888 | 0.834 | 0.888 | 0.832 | 0.860 | 0.828 | | |
| $U_{0.3,6}$ | 500 | 500 | 0.846 | 0.792 | 0.838 | 0.806 | 0.816 | 0.760 | 0.027 | 0.036 |
| | | 9500 | 0.819 | 0.756 | 0.817 | 0.782 | 0.795 | 0.741 | | |
| $U_{0.3,12}$ | 500 | 500 | 0.786 | 0.656 | 0.810 | 0.728 | 0.790 | 0.693 | 0.011 | 0.020 |
| | | 9500 | 0.785 | 0.657 | 0.799 | 0.708 | 0.779 | 0.677 | | |

| | | | | | | | | | | | |
|---|---|---|---|---|---|---|---|---|---|---|---|
| $U_{0.3,25}$ | 500 | 500 | 0.788 | 0.653 | 0.806 | 0.687 | 0.768 | 0.656 | 0.016 | 0.024 |
| | | 9500 | 0.773 | 0.629 | 0.790 | 0.663 | 0.772 | 0.665 | | |
| $U_{0.5,1}$ | 500 | 500 | 0.886 | 0.884 | 0.882 | 0.884 | 0.866 | 0.866 | 0.029 | 0.033 |
| | | 9500 | 0.857 | 0.853 | 0.854 | 0.851 | 0.844 | 0.843 | | |
| $U_{0.5,6}$ | 500 | 500 | 0.718 | 0.717 | 0.694 | 0.694 | 0.684 | 0.684 | 0.003 | 0.002 |
| | | 9500 | 0.721 | 0.719 | 0.719 | 0.718 | 0.700 | 0.700 | | |
| $U_{0.5,12}$ | 500 | 500 | 0.706 | 0.704 | 0.694 | 0.695 | 0.664 | 0.664 | 0.021 | 0.023 |
| | | 9500 | 0.685 | 0.681 | 0.678 | 0.676 | 0.643 | 0.642 | | |
| $U_{0.5,25}$ | 500 | 500 | 0.664 | 0.660 | 0.666 | 0.664 | 0.656 | 0.651 | 0.017 | 0.020 |
| | | 9500 | 0.664 | 0.655 | 0.817 | 0.662 | 0.639 | 0.631 | | |

## 7.3 The predicted probability of failure, $p_f$

In previous sections, the model performances on classifying the slopes as failed or stable were examined. The nature of binary classification results may induce a high number of near-to-failure samples being incorrectly classified reducing the classifier accuracies. However, in the case of the predictions of FOS data, slight errors in the prediction of the near-to-failure cases (FOS≈1) can induce a small error in the mean FOS calculated. Similarly, when calculating the $p_f$, each sample is predicted with a probability of being classified as unstable, and the mean of these probabilities is the predicted $p_f$ for the dataset. Therefore, the errors associated with the near-to-failure cases can be averaged and minimised in the calculated $p_f$ resulting in highly accurate predictions in the reliability analysis.

## 7.3.1 The effects of the size of training datasets on the predicted $p_f$

The number of MC simulations adequate for the training dataset to achieve accurate results is a critical point in machine learning aided MC analysis. In this study, for each case of statistical parameters (specific $COV$, $\xi$ and $\mu_{C_u}$), a full MC dataset consists of 2000 simulations. The ML models are progressively trained on different numbers of randomly selected samples to discover how the ML model can be improved by having larger portions of MC simulations as the training datasets. For each number of samples, the model training step is repeated 10 times and the mean of the results are reported. This study can help identify the optimum number of MC simulations required to efficiently achieve the expected accuracies when using ML surrogate models.

As shown in Fig. 6-a and b, accessing only a few MC samples as low as 10 (equal to 0.5% of MC data here) will be enough to limit the $p_f$ prediction errors in the order of 1%. It is noteworthy that the selected simulations should include enough examples from both failed and stable slopes particularly in the case of unbalanced classes. With increasing the number of training samples to 40 (2% of MC data), the model prediction performance improves and $p_{f_{err}}$ can be close to 1% ($p_{f_{err}} \approx \%1$). With the training dataset size reaching 400 samples (20% of MC data), the $p_{f_{err}}$ can decrease for one order of magnitude ($p_{f_{err}} \approx \%0.1$). Therefore, it may be concluded that for the studies where $p_f$ prediction errors limited to 1% is acceptable, it is sufficient to access 1-2% of MC samples to train the ML surrogate models.

### 7.3.2 The effects of the source of training datasets on the predicted $p_f$

The effects of the source of training samples on the efficiency of predictions are also examined. For the $p_f$ prediction on a dataset with given $COV$, $\xi$ and $\mu_{C_u}$ containing 2000 samples, two different training sources are considered: one is the same stochastically specific dataset; the other is stochastically diverse datasets with variable $COV$, $\xi$ and $\mu_{C_u}$ values (the entire of 120,000 samples). This comparison will demonstrate if a *unique* model trained on diverse datasets is appropriate for efficient predictions on various datasets or it is more efficient to train an individual model for each dataset using samples of the same dataset. As evident in Fig. 6, training the ML models on samples selected from the specific dataset to predict the results of the same dataset is absolutely more efficient than training a unique model on diverse datasets. Furthermore, the efficiency of the unique model trained on diverse datasets does not easily improve with the increased size of the training dataset. In contrast, the individual model trained on samples from the same specific dataset improves considerably by the increase of the training dataset size with more than two orders of magnitude smaller $p_{f_{err}}$ than that of the unique model for 500 training samples.

From the results shown in this section, it can be concluded that training several models each specific to a statistically different dataset is much more efficient than training a unique model on all datasets.

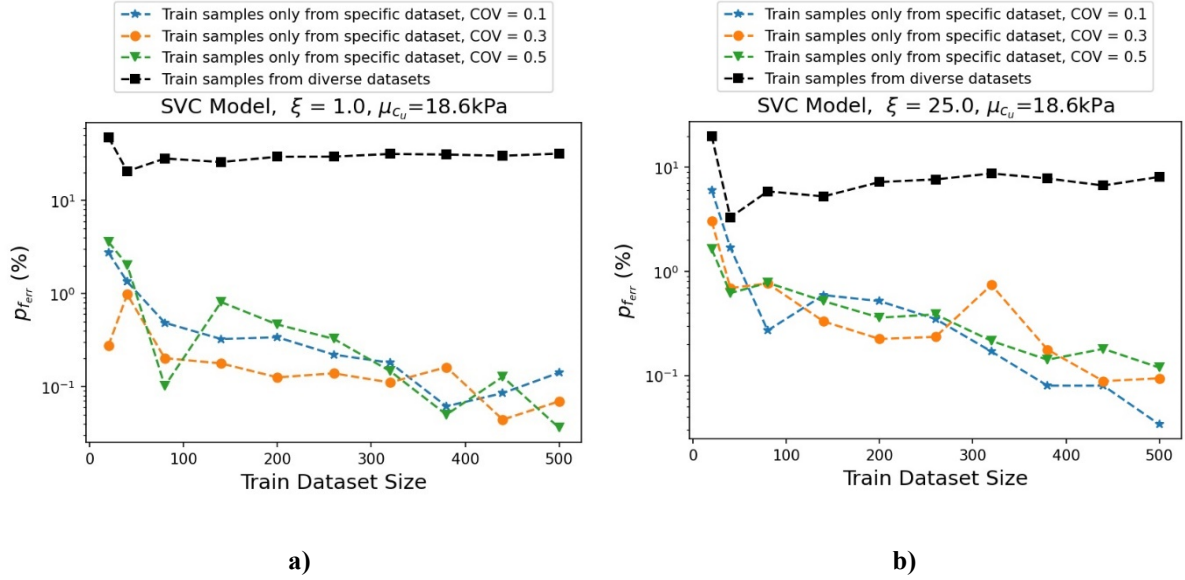

**Fig. 6** The errors in the ML predicted probability of failure, $p_{f_{err}}$ versus the number of samples in the training dataset. Each specific MC dataset (specific $COV$, $\xi$ and $\mu_{c_u}$) used in a) and b) contains 2000 samples. Diverse datasets consist of 120,000 samples from different $COV, \xi$ and $\mu_{c_u}$ values.

### 7.4 The accuracy of the predicted $p_f$

The ultimate results of MC slope stability analysis are the probability of failure. To confirm the efficiency of ML surrogate models in MLAMC reliability analysis of slopes, the errors in the probability of failure ($p_{f_{err}}$) obtained from the ML surrogate models are evaluated. In Fig. 7, the estimated $p_{f_{err}}$ is shown when using 1% and 2% of MC data with and without the application of ML surrogate models. The errors are calculated in comparison with actual $p_f$ values extracted from full-size MC data. When accessing only 1% of MC data (here as 20 samples out of 2000 for each dataset), the mean error of estimated $p_f$ (without any surrogate model) is 6.4%. This value means that if only 20 slope stability samples are generated instead of 2000 for each statistical case, the obtained $p_f$ may include an error of $\pm 6.4\%$ on average. However, with training ML surrogate models on the same 1% of MC data, the $p_{f_{err}}$ can reduce to 0.71% which is one order of magnitude more accurate (Fig. 7 a).

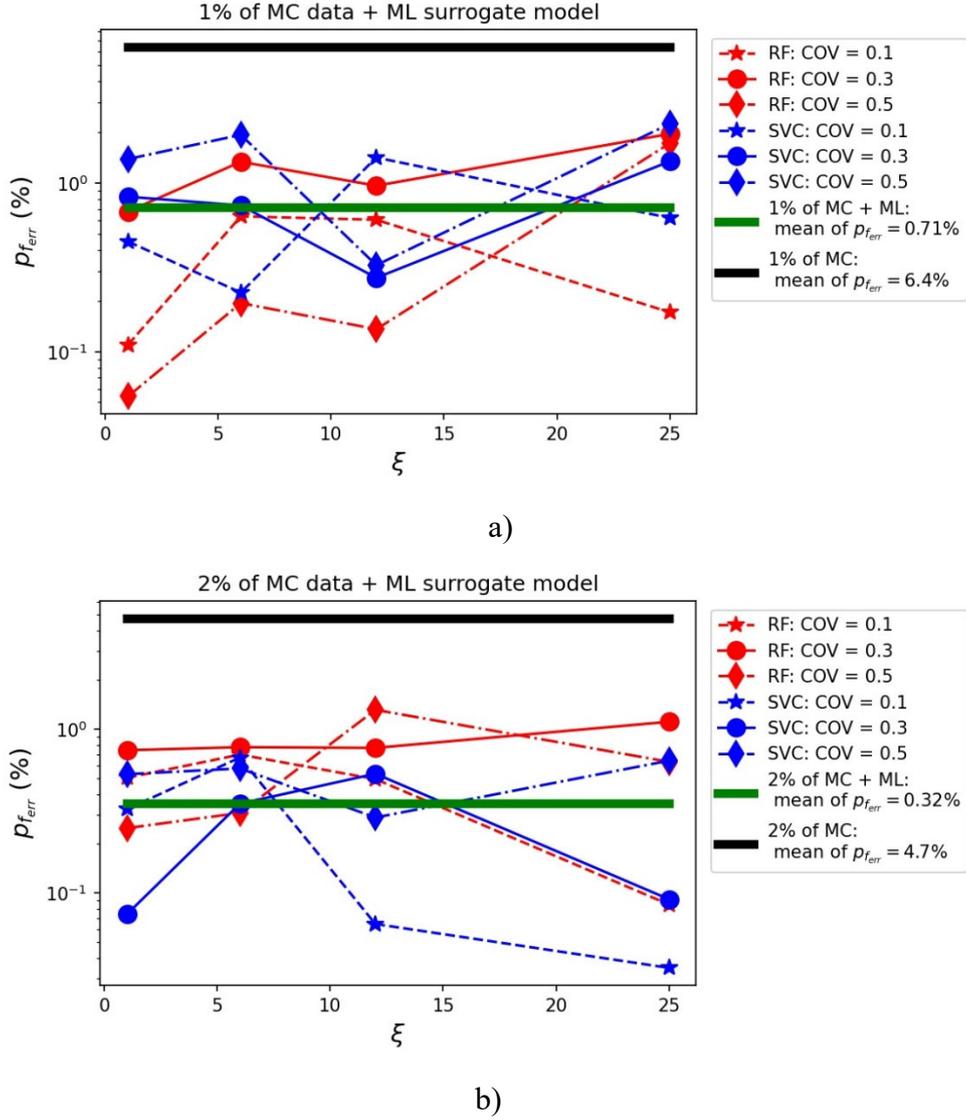

**Fig. 7** Errors in the ML predicted probability of failure ($p_{f_{err}}$) using only 1% (a) or 2% (b) of MC data (20 or 40 samples out of 2000). The solid black lines indicate the average of errors from the same amount of data without the application of ML surrogate models.

With using 2% of MC data (40 samples out of 2000 for each MC dataset) without any surrogate model, the $p_{f_{err}}$ is 4.7% on average (Fig. 7 b). However, when the same amount of data is used to train ML surrogate models, the predicted $p_f$ using ML models can be remarkably more accurate. The $p_{f_{err}}$ can decrease to only 0.32% on average which shows more than one order of magnitude improvement in the accuracy.

Fig. 7 also indicates that both RF and SVC models can perform similarly well on all datasets of different heterogeneity and anisotropy levels where the $p_{f_{err}}$ is never more than 2% when using 1 or 2% of MC data.

The results are also demonstrated as $p_f$ versus anisotropy ratio $\xi$ in Fig. 8 where the actual $p_f$ values obtained from complete MC datasets are shown as solid black lines. The shadow bonds in left panels are the mean errors when using only 1% (Fig. 8-a-left panel) or 2% (Fig. 8-b-left panel) of MC data equal to 20 and 40 slope stability samples from each dataset of 2000 samples. As depicted in Fig. 8-left panels, generating only 1 or 2% of MC data is obviously insufficient in obtaining acceptable $p_f$ results with mean errors exceeding the difference between the results of various $COV$ datasets. However, when the same amount of data is aided with ML surrogate models, the error margins are significantly reduced as shown in the corresponding plots at Fig. 8- right panels resulting in practically accurate predictions.

Fig. 9 provides a comparison between the $p_f$ resulted from full-size MC simulations (solid lines in left panels) and predicted by machine learning aided MC analysis using only 1% of data (20 samples out of each 2000 simulations for each $COV-\xi-\mu_{C_u}$ dataset) shown as dashed lines. The right panels of Fig. 9 also provide the errors in the estimated $p_f$ values by MLAMC analyses.

For $COV = 0.1$ (Fig. 9-a), the smallest value of $\mu_{C_u}$ equal to 18.6 kPa corresponding to $FOS_{det} = 1$ is associated with the largest $p_f$ values. Higher $\mu_{C_u}$ values with low heterogeneity ($COV = 0.1$) result in small $p_f$ approaching zero. The error in estimating the $p_f$ values using MLAMC analyses (1% of MC data aided by ML surrogate models) is always less than the error in the data-only results using 1% of MC data. The errors in the data-only results are about 9% on average (solid black line in Fig. 9-a-right panel) while the ML aided results are more than two orders of magnitude more accurate in many cases.

With a higher heterogeneity ($COV = 0.3$) shown in Fig. 9-b, the $p_f$ values will generally increase for different $\mu_{C_u}$ cases. The error in MLAMC estimated $p_f$ values using 1% of MC data is always less than the data-only errors (Fig. 9-b: right panel). The errors of ML aided $p_f$ results are below 1% in many cases with one to two orders of magnitude higher accuracy than the data-only results (Fig. 9-b: right panel).

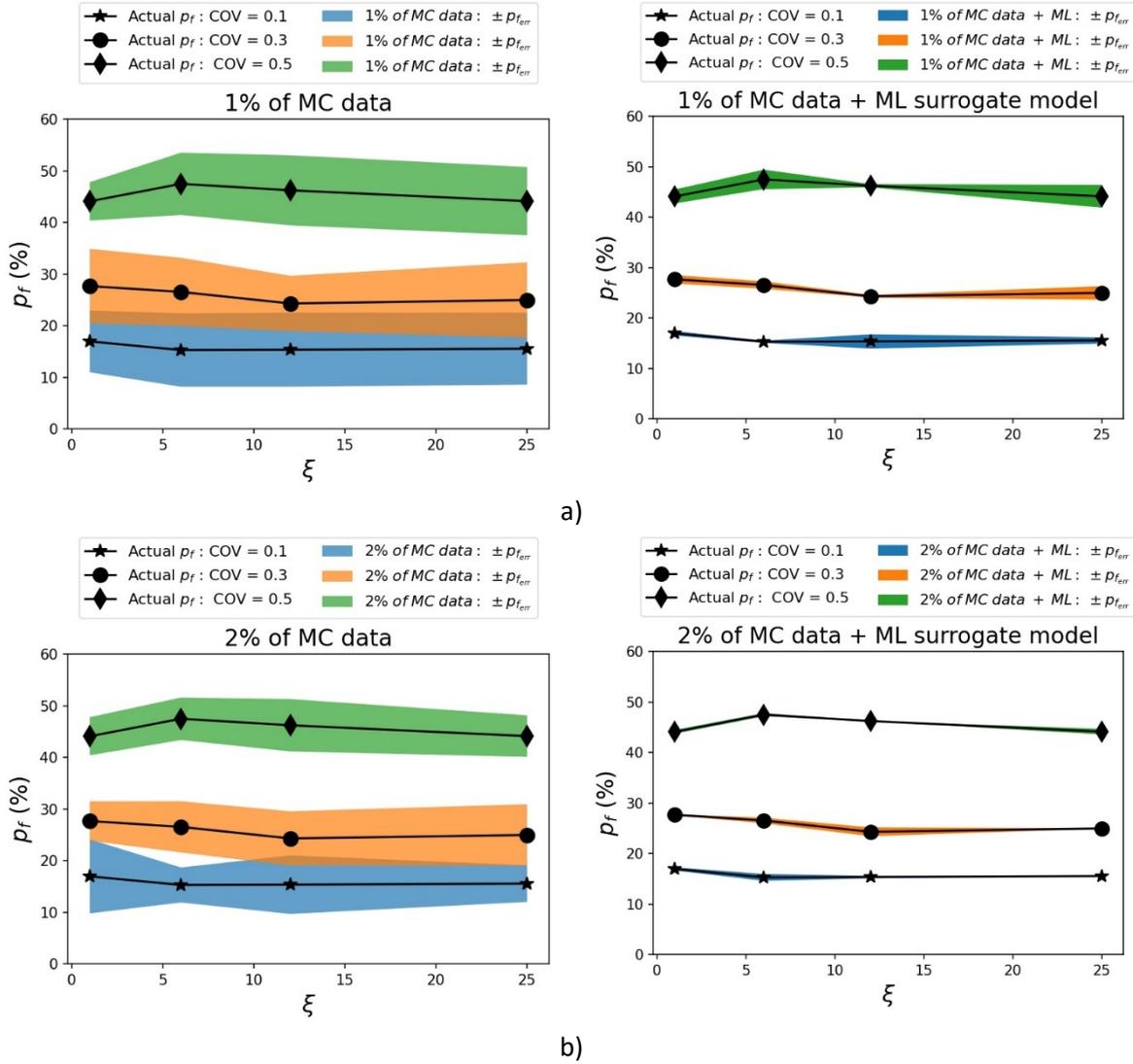

**Fig. 8** Error margins in the obtained probability of failure ($p_f$) from only 1% (a) or 2% (b) of MC data (left panels) versus errors in the ML aided predictions of $p_f$ using the same amount of data (right panels).

With the highest heterogeneity levels ($COV = 0.5$) as depicted in Fig. 9-c: left panel, the $p_f$ values increase with $\xi$ for relatively stronger soils ($\mu_{C_u} = 29.7$ - $33.5$ kPa) but decrease with $\xi$ for relatively weaker soils ($\mu_{C_u} = 18.6 - 22.3$ kPa). As shown in the right panel of Fig. 9-c, the errors in estimated $p_f$ values using 1% of MC data aided by ML surrogate models are always below the data-only errors with up to two orders of magnitude accuracy improvement.

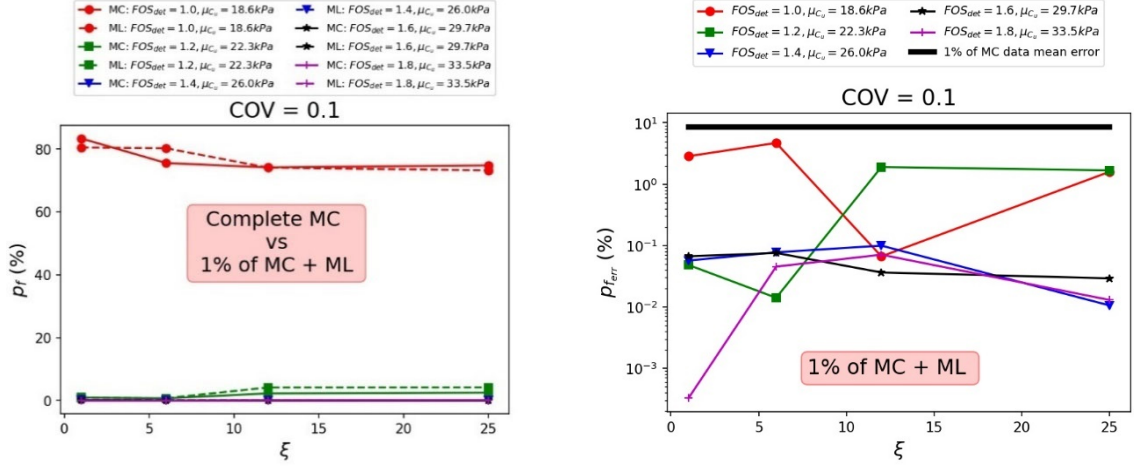

a) $COV = 0.1$

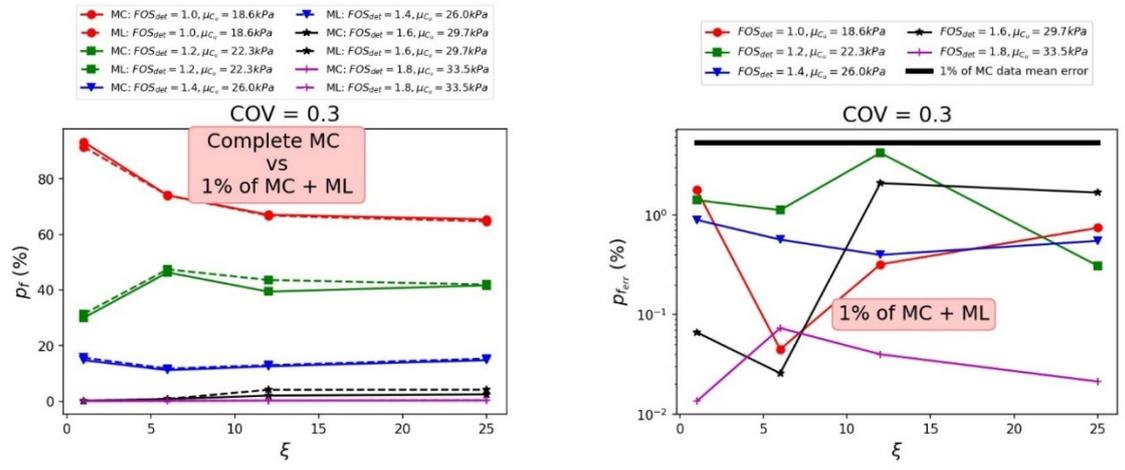

b) $COV = 0.3$

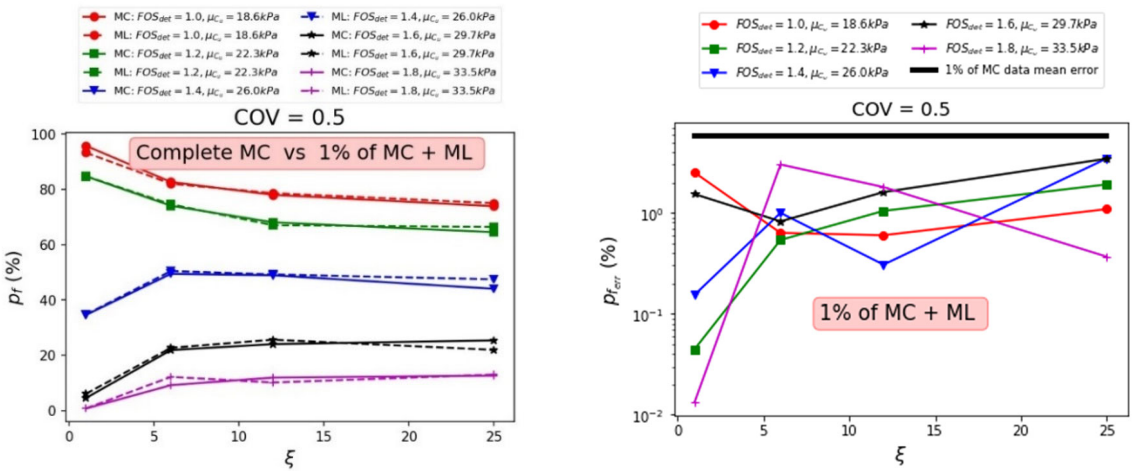

c) $COV = 0.5$

**Fig. 9** Probability of failure, $p_f$ obtained from complete Monte Carlo (MC) data versus Machine Learning (ML) aided Monte Carlo using only 1% of MC data (left panels), and the error in the estimated $p_f$ using ML surrogate models (right panels) for $COV = 0.1$ (a), $COV = 0.3$ (b) and $COV = 0.5$ (c). $p_f$ values are shown against the anisotropy ratio, $\xi$. The straight black lines in the right panels show the mean errors of data-only results (1% of MC data without ML surrogate models).

The results are also demonstrated in the form of $p_f$ versus the deterministic factor of safety, $FOS_{det}$ as shown in Fig. 10 left panels. In general, a higher $FOS_{det}$ corresponds to a relatively stronger soil leading to smaller $p_f$ values (larger reliability values). With larger spatial variations in soil strength parameters (i.e., larger $COV$), the impact of the anisotropy ratio $\xi$ can be more significant. In general, it can be concluded that a higher $\xi$ will decrease the $p_f$ for relatively weaker soils (e.g., $\mu_{C_u} = 18.6$ kPa) but will increase $p_f$ for relatively stronger soils (e.g., $\mu_{C_u} = 33.5$ kPa).

The errors in estimated $p_f$ values using 1% of MC data aided by ML models are normally below 1% ($p_f \lesssim 1\%$) and always less than data-only errors for $COV = 0.1 - 0.5$ (Fig. 10-right panels). The $p_f$ predictions using MLAMC analysis with 1% of MC data is always more accurate than data-only estimates normally for one to several orders of magnitude.

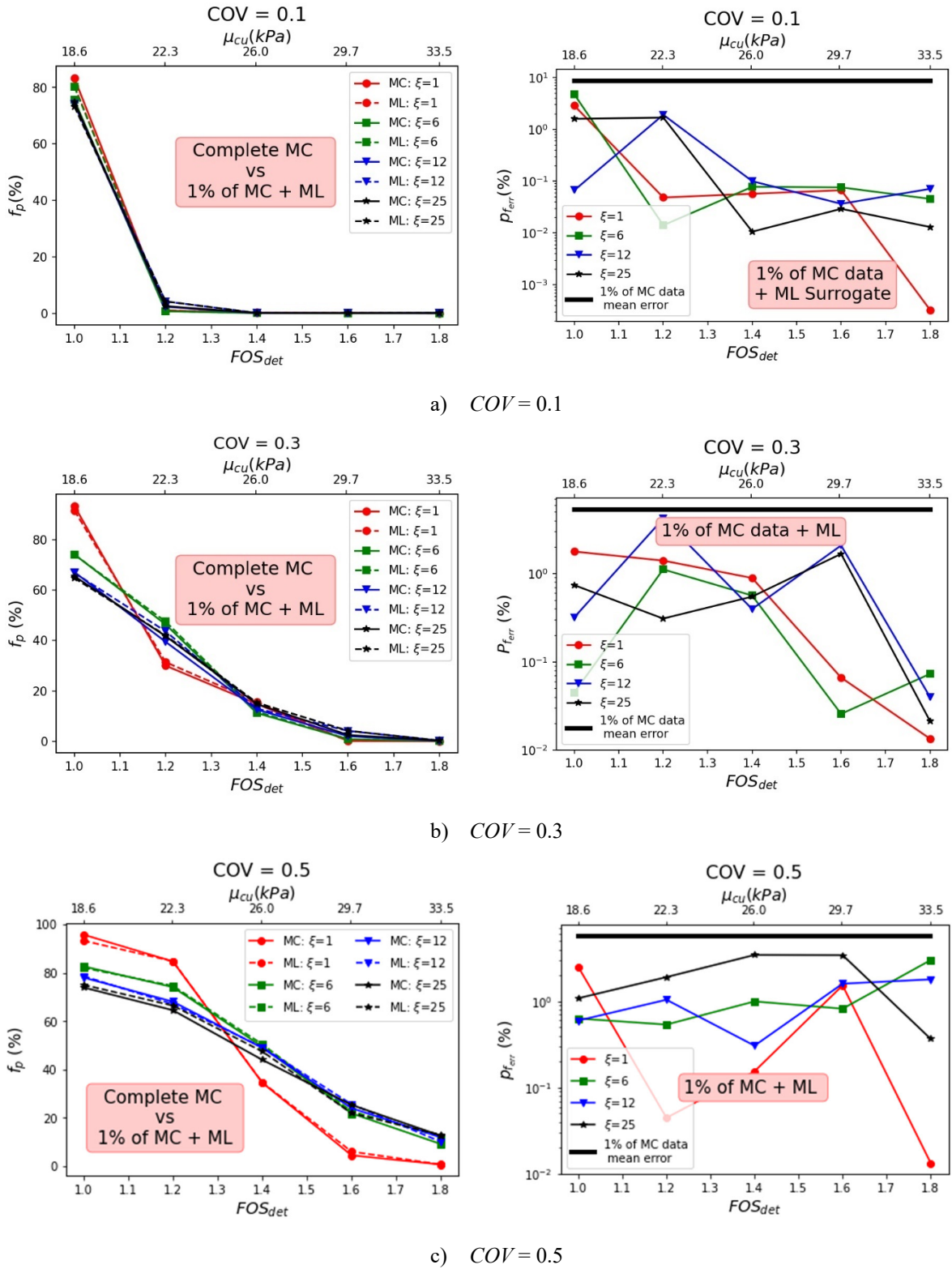

**Fig. 10** Probability of failure, $p_f$ obtained from Complete Monte Carlo data versus Machine Learning (ML) aided Monte Carlo using only 1% of MC data (left panels), and the error in the estimated $p_f$ using ML (right panels) for $COV = 0.1$ (a), $COV = 0.3$ (b) and $COV = 0.3$ (c). $p_f$ values are shown against the deterministic FOS and the mean of undrained shear strength, $\mu_{cu}$. The straight black lines in the right panels show the mean errors of data-only results (1% of MC data without ML surrogate models).

## 7.5. Computational time

The main challenge of the stochastic Monte Carlo modelling in geomechanics is the large computational time and power required. In this study, the CPU time for each random finite difference simulation to determine the slope stability status without a FOS calculation is 43 seconds on average on a Laptop with Intel Core i3-5010U 2.1GHz processor using four cores. Therefore, each dataset of particular ($COV, \xi, \mu_{C_u}$) values consisting of 2,000 simulations takes a CPU time of 23.9 hours to finish. The entire random field MC dataset containing 120,000 simulations with no FOS calculations is completed in about 60 days. In the case of the random finite difference simulations with FOS calculation using the strength reduction technique, the CPU time for each simulation can reach 220 seconds on average. Such a dataset with calculated FOSs will require up to 306 days to finish demanding unaffordable computational costs for such a study.

The ML surrogate models are trained on 1 and 2% of data (1200 and 2400 samples out of 120,000). The computational time for generating this training dataset will take only about 14 and 28 hours where the ML training and predictions are performed within a few minutes. Thus, the proposed ML surrogate models trained on samples with no FOS calculations can reduce the computational CPU time of such study from 306 days to only about 0.6 day, where a mean error of the predicted probabilities of failure can be about 0.7%. The mean error can decrease to about 0.3% for a computational time of 1.2 days. A summary of the computational time and expected errors are presented in Table 7.

**Table 7.** Comparison of the computational time of original and ML aided MC methods

| | Complete MC Data | | | 1% of MC data | | | | 2% of MC data | | |
|---|---|---|---|---|---|---|---|---|---|---|
| | | with FOS calculations | without FOS calculations | | | Data Only | Data + ML surrogate | | Data Only | Data + ML surrogate |
| **MC dataset** | MC samples | CPU time (hours) | CPU time (hours) | MC samples | CPU time (hours) | $P_{f_{err}}$ (%) | $P_{f_{err}}$ (%) | MC samples | CPU time (hours) | $P_{f_{err}}$ (%) | $P_{f_{err}}$ (%) |
| Specific datasets ($COV$, $\xi$, $\mu_{c_u}$) | 2,000 | 122.2 (~5.1 days) | 23.9 (~1 day) | 20 | 0.24 (~14 min) | Up to 9.4 | 0.05 - 2 | 40 | 0.48 (~28 min) | Up to 8.2 | 0.04 – 0.7 |
| Entire dataset | 120,000 | 7333 (~306 days) | 1433 (~60 days) | 1200 | 14.33 (~0.6 days) | 6.4 | 0.7 | 2400 | 28.66 (~1.2 days) | 4.7 | 0.32 |

## 8. Conclusion and future research direction

In this study, we investigated the efficiency of ML methods as response/surrogate models to predict the reliability results of the stochastic slope stability analysis. We compared a complete Monte Carlo reliability analysis of heterogeneous anisotropic slopes with the results of the machine learning aided MC reliability analysis using only 1-2% of the MC data. In the original MC study, random field finite difference simulations were conducted to examine the reliability of slopes. The undrained shear strength of soil is spatially variable within random fields with the mean value, $\mu_{c_u}$ varying in the range of 18.6 – 33.5 kPa, $COV$ ranging from 0.1 to 0.5, and the anisotropy ratio, $\xi$ ranging from 1 to 25. The outcome of MC simulations is only the failure or non-failure status of slopes circumventing the time-consuming calculation of FOS. In parallel, we implemented a Machine Learning aided Monte Carlo analysis with ML surrogate models trained on only 1-2% of random field data to predict the rest of the samples. With MLAMC analysis compared to the complete MC study, a realistic examination of the efficiency of ML surrogate models for the stochastic reliability analysis became possible. The study showed that the proposed method can reduce the computational time to 0.2% of the original MC study where the mean error in the predicted $p_f$ is limited to 0.7%.

The main findings of this study can be summarised as follows:

- Among RF, SVC and ANN classifiers, SVC outperformed others for the slope stability classification task on all random filed datasets. However, the SVC model is more sensitive to hyperparameters with up to 70% variations of accuracy in the tuning process. RF is the least sensitive model to tuning, with only 2.5% variations in accuracy. ANNs appear to be less efficient than other models, although after extensive optimisations.

- The performance of ML models in classifying slope stability status depends on the level of heterogeneity and anisotropy. The best performance (ACC = 0.962 and AUC = 0.983) are associated with the lowest heterogeneity ($COV$ = 0.1) and anisotropy ($\xi$ = 1). For the most heterogeneous and anisotropic slopes examined in this study ($COV$ = 0.5, $\xi$ = 25), the accuracy and AUC decreased to 0.691 and 0.727.

- *The size of testing datasets:* Testing ML surrogate models against randomly selected 5% of MC samples is statistically equivalent to testing against entire MC data (the error in classification accuracy estimations is less than 3%). Thus, accessing a small portion of the MC dataset to test the models is adequate to accurately evaluate the performance of ML models.

- *The size of training datasets:* Accessing only 20-40 samples (equal to 1-2% of MC datasets) to train the ML surrogate models can limit the $p_f$ prediction errors to $p_{f_{err}} \approx$ %1. With the training dataset size reaching 400 samples (20% of MC datasets), the $p_{f_{err}}$ can decrease for one order of magnitude ($p_{f_{err}} \approx$ %0.1). The selected simulations should include enough examples from both failed and stable slopes particularly in the case of unbalanced classes.

- Training several models each specific to a statistically different dataset (specific $COV$, $\xi$ and $\mu_{C_u}$) is much more efficient than training a unique model on all datasets.

- A $p_f$ calculated from 2% and 1% of MC data (without surrogate models) can incur about 4.7-6.4% of errors on average. Using the same amount of data, ML surrogate models can reduce $p_f$ errors to 0.32-0.71% on average, which is a one-order-of-magnitude accuracy improvement.

- For all heterogeneity and anisotropy levels examined, ML surrogate models trained on 1% of MC data consistently provide accurate $p_f$ results with errors in the range of 1-$10^{-2}$%.

- Circumventing the calculation of FOS for each sample and using the ML surrogate models, the MLAMC reliability method proposed in this study can reduce the computational time required for a stochastic study from 306 days to only 14 hours.

To further investigate the application of ML algorithms as surrogate models for MC reliability analysis, research can be extended using various random field MC data on different problems such as bearing capacity, groundwater modelling, settlement, deep foundations, retaining walls and liquefaction. Feature classification to identify geometrical features such as anisotropy attributes (layering effects) can also be a potential element to improve predictions. Further investigations on the training of deep and convolutional neural networks on random field data with failure and non-failure results (without FOS calculations) can also further confirm the potential of the machine learning approaches as a computationally efficient stochastic reliability analysis.

**Acknowledgement**

This research is funded by the Australian Research Council via the Discovery Projects (No. DP200100549).

**Data Availability Statement**

All data that support the findings of this study are available from the corresponding author upon reasonable request.

**References:**

[1] G.A. Fenton, D.V. Griffiths, Risk assessment in geotechnical engineering, John Wiley & Sons New York2008.

[2] J.T. Christian, C.C. Ladd, G.B. Baecher, Reliability applied to slope stability analysis, Journal of Geotechnical Engineering 120(12) (1994) 2180-2207.

[3] J. Ching, K.-K. Phoon, Probability distribution for mobilised shear strengths of spatially variable soils under uniform stress states, Georisk: Assessment and Management of Risk for Engineered Systems and Geohazards 7(3) (2013) 209-224.

[4] M. Lloret-Cabot, G.A. Fenton, M.A. Hicks, On the estimation of scale of fluctuation in geostatistics, Georisk: Assessment and management of risk for engineered systems and geohazards 8(2) (2014) 129-140.

[5] E. Vanmarcke, Random fields: analysis and synthesis, World scientific2010.

[6] J. Pieczyńska-Kozłowska, W. Puła, D. Griffiths, G. Fenton, Influence of embedment, self-weight and anisotropy on bearing capacity reliability using the random finite element method, Computers and Geotechnics 67 (2015) 229-238.

[7] J. Huang, D. Griffiths, G.A. Fenton, Probabilistic analysis of coupled soil consolidation, Journal of Geotechnical and Geoenvironmental Engineering 136(3) (2010) 417-430.

[8] D. Griffiths, G.A. Fenton, Influence of soil strength spatial variability on the stability of an undrained clay slope by finite elements, Slope stability 20002000, pp. 184-193.

[9] D. Griffiths, G.A. Fenton, Probabilistic slope stability analysis by finite elements, Journal of geotechnical and geoenvironmental engineering 130(5) (2004) 507-518.

[10] Y. Wang, Z. Cao, S.-K. Au, Efficient Monte Carlo simulation of parameter sensitivity in probabilistic slope stability analysis, Computers and Geotechnics 37(7-8) (2010) 1015-1022.

[11] Y. Wang, Z. Cao, S.-K. Au, Practical reliability analysis of slope stability by advanced Monte Carlo simulations in a spreadsheet, Canadian Geotechnical Journal 48(1) (2011) 162-172.

[12] E.H. Vanmarcke, Probabilistic modeling of soil profiles, Journal of the geotechnical engineering division 103(11) (1977) 1227-1246.

[13] E.H. Vanmarcke, Reliability of earth slopes, Journal of the Geotechnical Engineering Division 103(11) (1977) 1247-1265.

[14] J. Huang, G. Fenton, D.V. Griffiths, D. Li, C. Zhou, On the efficient estimation of small failure probability in slopes, Landslides 14(2) (2017) 491-498.

[15] F. Chen, L. Wang, W. Zhang, Reliability assessment on stability of tunnelling perpendicularly beneath an existing tunnel considering spatial variabilities of rock mass properties, Tunnelling and Underground Space Technology 88 (2019) 276-289.

[16] S.E. Cho, Effects of spatial variability of soil properties on slope stability, Engineering Geology 92(3-4) (2007) 97-109.

[17] J. Li, Y. Tian, M.J. Cassidy, Failure mechanism and bearing capacity of footings buried at various depths in spatially random soil, Journal of Geotechnical and Geoenvironmental Engineering 141(2) (2015) 04014099.

[18] S.-K. Au, J. Beck, Important sampling in high dimensions, Structural safety 25(2) (2003) 139-163.

[19] S.-K. Au, J.L. Beck, Estimation of small failure probabilities in high dimensions by subset simulation, Probabilistic engineering mechanics 16(4) (2001) 263-277.

[20] I. Papaioannou, W. Betz, K. Zwirglmaier, D. Straub, MCMC algorithms for subset simulation, Probabilistic Engineering Mechanics 41 (2015) 89-103.

[21] S.-H. Jiang, J.-S. Huang, Efficient slope reliability analysis at low-probability levels in spatially variable soils, Computers and Geotechnics 75 (2016) 18-27.

[22] Z.-P. Deng, M. Pan, J.-T. Niu, S.-H. Jiang, W.-W. Qian, Slope reliability analysis in spatially variable soils using sliced inverse regression-based multivariate adaptive regression spline, Bulletin of Engineering Geology and the Environment 80(9) (2021) 7213-7226.

[23] S.-H. Jiang, D.-Q. Li, L.-M. Zhang, C.-B. Zhou, Slope reliability analysis considering spatially variable shear strength parameters using a non-intrusive stochastic finite element method, Engineering geology 168 (2014) 120-128.

[24] F. Kang, Q. Xu, J. Li, Slope reliability analysis using surrogate models via new support vector machines with swarm intelligence, Applied Mathematical Modelling 40(11-12) (2016) 6105-6120.

[25] L. Liu, S. Zhang, Y.-M. Cheng, L. Liang, Advanced reliability analysis of slopes in spatially variable soils using multivariate adaptive regression splines, Geoscience Frontiers 10(2) (2019) 671-682.

[26] L.-L. Liu, Y.-M. Cheng, System reliability analysis of soil slopes using an advanced kriging metamodel and quasi–Monte Carlo simulation, International Journal of Geomechanics 18(8) (2018) 06018019.

[27] L. Wang, C. Wu, X. Gu, H. Liu, G. Mei, W. Zhang, Probabilistic stability analysis of earth dam slope under transient seepage using multivariate adaptive regression splines, Bulletin of Engineering Geology and the Environment 79(6) (2020) 2763-2775.

[28] P. Zeng, T. Zhang, T. Li, R. Jimenez, J. Zhang, X. Sun, Binary classification method for efficient and accurate system reliability analyses of layered soil slopes, Georisk: Assessment and Management of Risk for Engineered Systems and Geohazards (2020) 1-17.

[29] J. Huang, G. Fenton, D. Griffiths, D. Li, C. Zhou, On the efficient estimation of small failure probability in slopes, Landslides 14(2) (2017) 491-498.

[30] X. He, F. Wang, W. Li, D. Sheng, Deep learning for efficient stochastic analysis with spatial variability, Acta Geotechnica (2021).

[31] X. He, F. Wang, W. Li, D. Sheng, Efficient reliability analysis considering uncertainty in random field parameters: Trained neural networks as surrogate models, Computers and Geotechnics 136 (2021) 104212.

[32] X. He, H. Xu, H. Sabetamal, D. Sheng, Machine learning aided stochastic reliability analysis of spatially variable slopes, Computers and Geotechnics 126 (2020) 103711.

[33] Z.Z. Wang, C. Xiao, S.H. Goh, M.-X. Deng, Metamodel-Based Reliability Analysis in Spatially Variable Soils Using Convolutional Neural Networks, Journal of Geotechnical and Geoenvironmental Engineering 147(3) (2021) 04021003.

[34] Z.-Z. Wang, S.H. Goh, Novel approach to efficient slope reliability analysis in spatially variable soils, Engineering Geology 281 (2021) 105989.

[35] K. Pham, D. Kim, S. Park, H. Choi, Ensemble learning-based classification models for slope stability analysis, Catena 196 (2021) 104886.

[36] C. Guardiani, E. Soranzo, W. Wu, Time-dependent reliability analysis of unsaturated slopes under rapid drawdown with intelligent surrogate models, Acta Geotechnica (2021) 1-26.

[37] S. Lin, H. Zheng, B. Han, Y. Li, C. Han, W. Li, Comparative performance of eight ensemble learning approaches for the development of models of slope stability prediction, Acta Geotechnica (2022).

[38] Z.Z. Wang, S.H. Goh, A maximum entropy method using fractional moments and deep learning for geotechnical reliability analysis, Acta Geotechnica (2021).

[39] L. Wang, C. Wu, L. Tang, W. Zhang, S. Lacasse, H. Liu, L. Gao, Efficient reliability analysis of earth dam slope stability using extreme gradient boosting method, Acta Geotechnica 15(11) (2020) 3135-3150.

[40] J.P. Bharti, P. Mishra, V. Sathishkumar, Y. Cho, P. Samui, Slope Stability Analysis Using Rf, Gbm, Cart, Bt and Xgboost, Geotechnical and Geological Engineering 39(5) (2021) 3741-3752.

[41] H. Zhang, H. Nguyen, X.-N. Bui, B. Pradhan, P.G. Asteris, R. Costache, J. Aryal, A generalized artificial intelligence model for estimating the friction angle of clays in evaluating slope stability using a deep neural network and Harris Hawks optimization algorithm, Engineering with Computers (2021) 1-14.

[42] M. Aminpour, R. Alaie, N. Kardani, S. Moridpour, M. Nazem, Slope stability predictions on spatially variable random fields using machine learning surrogate models, manuscript submitted for publication (2022).

[43] K.-K. Phoon, J.V. Retief, Reliability of geotechnical structures in ISO2394, CRC Press2016.

[44] M. Jaksa, W. Kaggwa, P. Brooker, Experimental evaluation of the scale of fluctuation of a stiff clay, Proc. 8th Int. Conf. on the Application of Statistics and Probability, Sydney, AA Balkema, Rotterdam, 1999, pp. 415-422.

[45] K.-K. Phoon, F.H. Kulhawy, Characterization of geotechnical variability, Canadian geotechnical journal 36(4) (1999) 612-624.

[46] F. Cafaro, C. Cherubini, Large sample spacing in evaluation of vertical strength variability of clayey soil, Journal of geotechnical and geoenvironmental engineering 128(7) (2002) 558-568.

[47] H. El-Ramly, N. Morgenstern, D. Cruden, Probabilistic stability analysis of a tailings dyke on presheared clay shale, Canadian Geotechnical Journal 40(1) (2003) 192-208.

[48] M. Uzielli, G. Vannucchi, K. Phoon, Random field characterisation of stress-nomalised cone penetration testing parameters, Geotechnique 55(1) (2005) 3-20.

[49] C.N. Liu, C.-H. Chen, Spatial correlation structures of CPT data in a liquefaction site, Engineering Geology 111(1-4) (2010) 43-50.

[50] R. Suchomel, D. Mašín, Probabilistic analyses of a strip footing on horizontally stratified sandy deposit using advanced constitutive model, Computers and Geotechnics 38(3) (2011) 363-374.

[51] E. Bombasaro, T. Kasper, Evaluation of spatial soil variability in the Pearl River Estuary using CPTU data, Soils and Foundations 56(3) (2016) 496-505.

[52] G.B. Baecher, J.T. Christian, Reliability and statistics in geotechnical engineering, John Wiley & Sons2005.

[53] D.V. Griffiths, G.A. Fenton, Probabilistic methods in geotechnical engineering, Springer Science & Business Media2007.

[54] W.T. Vetterling, W.H. Press, S.A. Teukolsky, B.P. Flannery, Numerical recipes example book (c++): The art of scientific computing, Cambridge University Press2002.

[55] H. Zhu, L.M. Zhang, Characterizing geotechnical anisotropic spatial variations using random field theory, Canadian Geotechnical Journal 50(7) (2013) 723-734.

[56] H. Zhu, L. Zhang, T. Xiao, X. Li, Generation of multivariate cross-correlated geotechnical random fields, Computers and Geotechnics 86 (2017) 95-107.

[57] A.I. El-Kadi, S.A. Williams, Generating two-dimensional fields of autocorrelated, normally distributed parameters by the matrix decomposition technique, Groundwater 38(4) (2000) 530-532.

[58] R. Popescu, G. Deodatis, A. Nobahar, Effects of random heterogeneity of soil properties on bearing capacity, Probabilistic Engineering Mechanics 20(4) (2005) 324-341.

[59] R. Jamshidi Chenari, R. Alaie, Effects of anisotropy in correlation structure on the stability of an undrained clay slope, Georisk: Assessment and Management of Risk for Engineered Systems and Geohazards 9(2) (2015) 109-123.

[60] W.S. Noble, What is a support vector machine?, Nature biotechnology 24(12) (2006) 1565-1567.

[61] C. Cortes, V. Vapnik, Support-vector networks, Machine learning 20(3) (1995) 273-297.

[62] M. Belgiu, L. Drăguţ, Random forest in remote sensing: A review of applications and future directions, ISPRS journal of photogrammetry and remote sensing 114 (2016) 24-31.

[63] N. Kardani, A. Bardhan, S. Gupta, P. Samui, M. Nazem, Y. Zhang, A. Zhou, Predicting permeability of tight carbonates using a hybrid machine learning approach of modified equilibrium optimizer and extreme learning machine, Acta Geotechnica (2021) 1-17.

[64] S. Dreiseitl, L. Ohno-Machado, Logistic regression and artificial neural network classification models: a methodology review, Journal of biomedical informatics 35(5-6) (2002) 352-359.

[65] S. Suman, S. Khan, S. Das, S. Chand, Slope stability analysis using artificial intelligence techniques, Natural Hazards 84(2) (2016) 727-748.